\documentclass{article}
\usepackage{iclr2027_conference,times}
\usepackage{makecell}
\usepackage{hyperref}
\usepackage{url}
\usepackage{graphicx}
\usepackage{amsmath}
\usepackage{amssymb}
\usepackage{booktabs}
\usepackage[table]{xcolor}
\usepackage{pgfplots}
\pgfplotsset{compat=1.18}
\usepackage{natbib}
\usepackage{caption}
\usepackage{pifont}
\usepackage{wrapfig}
\usepackage{fontawesome5}
\newcommand{\xmark}{\ding{55}}
\definecolor{groupblue}{RGB}{235,245,255}

\newcommand{\stageI}{\uppercase\expandafter{\romannumeral1}}
\newcommand{\stageII}{\uppercase\expandafter{\romannumeral2}}

\title{DIDO: Distilling Interaction-Centric Dynamics into One-Step
Denoising for World Action Models}

\iclrfinalcopy

\author{
\centering
\normalsize
\begin{tabular}[t]{c}
Jing Lyu$^{*,1,2,3}$ \quad
Shuanghao Bai$^{*,4}$ \quad
Runze Xiao$^{3}$ \quad
Zhenyu Liao$^{6}$ \quad
Wenxing Tan$^{3}$ \\[2pt]
Zihan Tang$^{5}$ \quad
Ruochuan Shi$^{1,2,3}$ \quad
Cheng Peng$^{1,2,3}$ \quad
Yuheng Ji$^{1,2,3}$ \quad
Yihao Wang$^{3}$ \\[2pt]
Badong Chen$^{4}$ \quad
Pengwei Wang$^{3}$ \quad
Zhongyuan Wang$^{\dagger,3}$ \quad
Xiaoguang Zhao$^{\dagger,1,2}$ \\[6pt]
\normalfont\small $^{1}$Institute of Automation, Chinese Academy of Sciences \\[2pt]
\normalfont\small $^{2}$School of Artificial Intelligence, University of Chinese Academy of Sciences \\[2pt]
\normalfont\small $^{3}$Beijing Academy of Artificial Intelligence (BAAI) \\[2pt]
\normalfont\small $^{4}$Institute of Artificial Intelligence and Robotics, Xi'an Jiaotong University \\[2pt]
\normalfont\small $^{5}$Tsinghua University \quad $^{6}$Amazon \\[4pt]
\normalfont\small $^{*}$Equal contribution. \quad $^{\dagger}$Corresponding authors. \\[4pt]
\hypersetup{urlcolor=blue}
\href{https://loveju1y.github.io/DIDO/}{\faGlobe\textbf{Project Page}}
\quad \href{https://github.com/LoveJu1y/DIDO-WAM/}{\faGithub\textbf{Code}}
\end{tabular}
}

\begin{document}

\maketitle

\lhead{Preprint.}

\begin{abstract}
World Action Models (WAMs) use video generation models to predict future visual dynamics for robotic manipulation, but iterative denoising introduces additional latency for closed-loop control. One-step truncation reduces this cost, but risks discarding predictive information formed during later denoising steps. We empirically find that visual content converges at different rates during denoising: static scene structure forms early, whereas gripper--object interaction dynamics emerge progressively through subsequent steps. Consequently, one-step truncation preserves scene structure but loses late-emerging interaction dynamics useful for manipulation. Guided by this observation, we propose DIDO, which distills the converged dynamics of a multi-step video model into a single denoising step. By recovering interaction dynamics that would otherwise be lost through truncation, DIDO accelerates inference without sacrificing control performance. Moreover, DIDO introduces interaction-centric representation guidance to establish finer-grained, entity-level understanding of the recovered dynamic regions. Compared with Fast-WAM-IDM, DIDO reduces end-to-end inference latency by 87.6\% (from 871 to 108\,ms) while achieving higher success rates across multiple benchmarks. In general, DIDO achieves average success rates of 99.0\% on LIBERO, 76.3\% on LIBERO-Plus, and 91.9\% on RoboTwin, while also demonstrating effective transfer to long-horizon and generalization tasks in real-world robotic manipulation.

\end{abstract}

\section{Introduction}

Vision-Language-Action Models (VLAs) have achieved strong performance in robotic manipulation by leveraging large-scale vision-language pretraining to follow language instructions across diverse tasks and embodiments~\citep{brohan2023rt1,zitkovich2023rt2,kim2025openvla,intelligence2025pi05}. Most VLA policies predict actions directly from the current observation and language instruction, without explicitly modeling how the environment will evolve under the robot's actions. Recently, World Action Models (WAMs) have explored a complementary direction by using video generation models to predict future visual dynamics~\citep{bai2026embodied,wu2024gr1,hu2025videopredictionpolicy,ye2026dreamzero}. These predictions provide the policy with explicit foresight into how the robot, manipulated objects, and their interactions evolve over time. However, realizing such foresight at inference time remains computationally expensive because video generation typically relies on iterative denoising.

A natural way to reduce this cost is to truncate the denoising process and extract predictive representations from only one or a few early steps~\citep{hu2025videopredictionpolicy,chi2025mind}. This substantially improves inference efficiency and has been shown to retain useful predictive information. Yet it remains unclear how much of the dynamics represented by the fully denoised video model are already present at these early steps. In particular, when denoising is reduced to a single step, what dynamics are preserved, and what dynamics are still missing?

\begin{figure*}[t]
\centering
\includegraphics[width=1\textwidth]{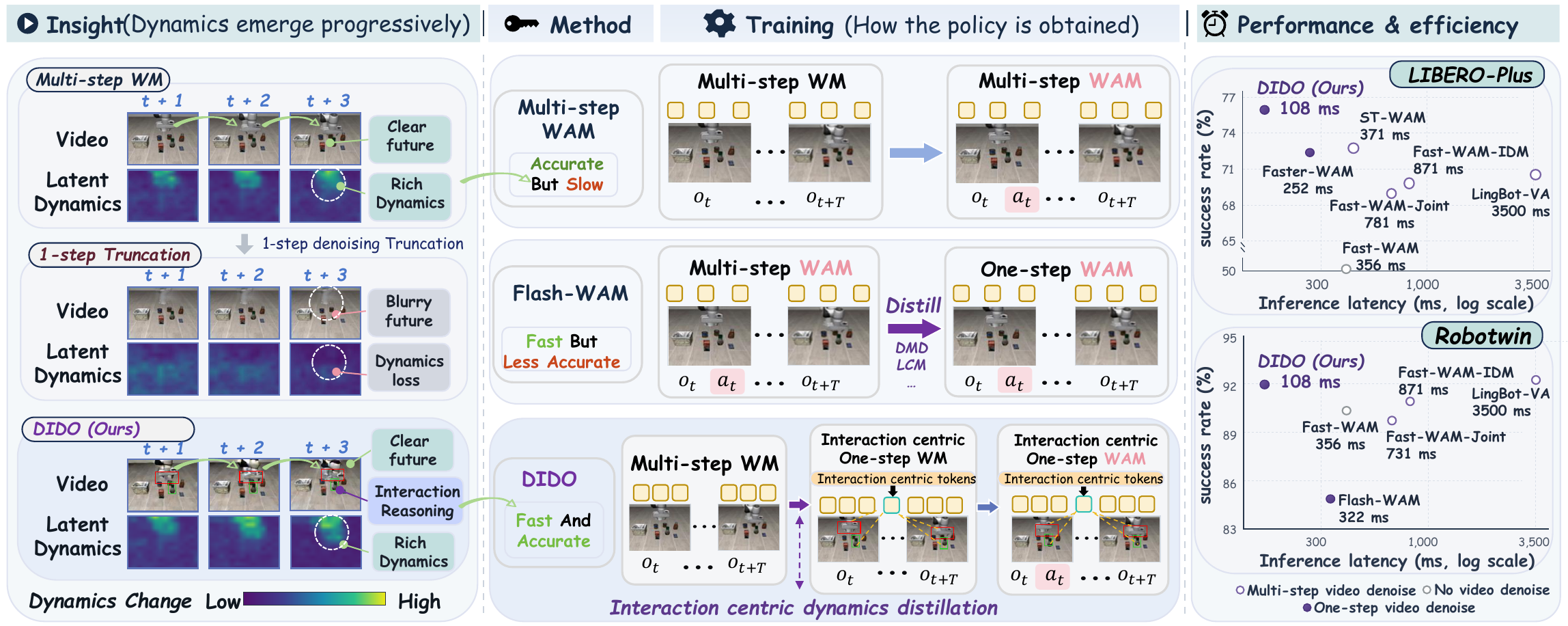}
\caption{
\textbf{DIDO: interaction-centric dynamics in one step.}
The insight panel contrasts multi-step prediction, one-step truncation,  and DIDO through decoded videos and latent dynamics, demonstrating that dynamics emerge progressively through subsequent steps and DIDO recover these dynamics in one step. The training panel summarizes interaction-centric distillation. The performance--efficiency panel compares success rates against inference latency on LIBERO-Plus and RoboTwin, highlighting DIDO's strong performance at 108\,ms.
}
\label{fig:trajectory}
\end{figure*}

To answer this question, we conduct an empirical study of the denoising trajectory in Figure~\ref{fig:trajectory}, examining both decoded videos and latent representations. We observe a clear spatial and temporal asymmetry. After a single step, static background structure is already well formed, whereas the gripper and manipulated object remain blurry and poorly resolved. The same pattern appears in latent space: inter-frame changes in these interaction regions are weak at early steps but progressively strengthen and become spatially localized as denoising proceeds. These changes are concentrated around the gripper and manipulated object, whose evolving spatial relationship provides predictive cues for action selection. Naive single-step truncation therefore largely preserves scene structure but misses the interaction-centric dynamics that emerge later in denoising. Our denoising-stage analysis further shows that conditioning the action expert on later representations improves manipulation success. Together, these findings highlight the importance of interaction-centric dynamics for manipulation.

We instantiate this insight in DIDO, a one-step World Action Model centered on recovering and better grounding interaction-centric dynamics. In Stage~I, DIDO jointly combines distribution matching distillation (DMD)~\citep{yin2024dmd2} with interaction-centric visual reasoning. DMD transfers the converged dynamics of a multi-step video model into a single denoising step. Beyond recovering these dynamics, interaction-centric visual reasoning provides finer-grained, entity-level representations of the dynamic regions. Interaction-centric tokens, guided by future bounding-box trajectories, explicitly model the gripper, target object, and their evolving spatial relationship. Multi-layer alignment with target-object features from a pretrained DINOv3 encoder~\citep{simeoni2026dinov3} preserves discriminative visual information to support this entity-level modeling. Distillation and interaction-centric visual reasoning are jointly optimized in this stage. In Stage~II, the resulting single-step world model is coupled with an action expert through a mixture-of-transformers architecture for closed-loop control. Overall, DIDO recovers interaction dynamics within a single denoising step, avoiding costly iterative denoising while maintaining or improving control performance. Ablations further show that interaction-centric guidance improves manipulation success, supporting the value of explicitly modeling these dynamic regions.

The main contributions are threefold:
\begin{enumerate}
\item We reveal that static scene structure forms early during video denoising, while manipulation-relevant dynamics around the gripper and target object emerge progressively at later steps, and connect this observation to the control benefits of later denoising representations.
\item Guided by this finding, we develop DIDO to recover late-emerging dynamics in one step and explicitly structure them around manipulation-relevant entities, achieving efficient dynamics modeling for control.
\item Rather than merely preserving performance under one-step compression, DIDO achieves higher manipulation success rates while substantially reducing inference latency. Compared with Fast-WAM-IDM, which has the same model size and a similar architecture, DIDO reduces end-to-end latency by 87.6\% (from 871 to 108\,ms) while improving success rates across multiple benchmarks. DIDO achieves 99.0\% on LIBERO, 76.3\% on LIBERO-Plus, and 91.9\% on RoboTwin, with effective transfer to real-world manipulation.
\end{enumerate}

\section{Related Work}

\paragraph{World Action Models for Robotic Manipulation.}
Vision-Language-Action Models (VLAs) have achieved strong performance in robotic manipulation~\citep{brohan2023rt1,zitkovich2023rt2,kim2025openvla,intelligence2025pi05,bai2026laravla}. World Action Models (WAMs) explore a complementary direction by incorporating video-based future modeling into action prediction~\citep{ye2026dreamzero,li2026lingbotva}. Early approaches synthesize future frames or subgoal images to guide action generation~\citep{du2023unipi,black2024susie,tian2025seer,wu2024gr1}, while subsequent methods condition policies on predictive representations ~\citep{hu2025videopredictionpolicy,wang2026orca,lyu2026lda1b}. Recent WAMs further couple video and action through joint generative modeling, latent action representations, or multi-stream architectures~\citep{cheang2024gr2,bi2026motus,team2026motubrain}. Action-centered variants also use future visual dynamics primarily as training supervision or optional inference-time guidance~\citep{ye2026gigaworld,team2026gigaworld05}. Across these designs, a central question is which aspects of future dynamics are most useful for control. DIDO addresses this question by focusing on gripper–object interaction dynamics that emerge during later denoising steps and recovering them in a single step.

\paragraph{Efficient World Action Models.}
Recent work improves WAM efficiency through several strategies. One line removes test-time future imagination entirely. Fast-WAM predicts actions without generating future observations~\citep{yuan2026fastwam}, eliminating the cost of explicit future prediction. A second line retains future modeling but truncates denoising, conditioning the policy on representations extracted from only one or a few early steps~\citep{chi2025mind,hu2025videopredictionpolicy,zhao2026fasterwam}. Other approaches preserve future prediction while reducing architectural or conditioning overhead, for example by simplifying the action module, reusing future features across action denoising steps~\citep{ma2026fasterwam}, or moving auxiliary geometric supervision to training time~\citep{zhang2026mecowam}. Closest to DIDO are methods that distill multi-step video or video-action predictions into one-step generation or predictive representations~\citep{akbari2026flashwam,yan2026svam}. Within this line of work, DIDO uses interaction-centric supervision to guide what the one-step model preserves during distillation. Jointly optimized with distribution matching, trajectory and visual supervision encourage the distilled representations to retain manipulation-relevant entity dynamics and their coupled evolution.

\paragraph{Object- and Interaction-Centric Representations for Manipulation.}
Object-centric VLA methods structure policy representations around task-relevant entities using visual prompts, object-conditioned features, slots and relations, or 3D object representations~\citep{li2025crayonrobo,li2025controlvla,hanyu2025slotvla,liu2026sam3d}. Recent WAMs extend this idea to future prediction by modeling object masks~\citep{yu2026maskwam,lou2026maskworldmodel}, temporally persistent robot/object slots~\citep{liu2026oawam}, or structured semantic, geometric, and kinematic foresight~\citep{yan2026svam,jin2026structvla,zheng2026emerging,ji2025robobrain}. While these approaches make future prediction increasingly object- and structure-aware, robot--object interactions are still largely represented implicitly. DIDO instead adopts interaction-centric visual reasoning, explicitly integrating gripper--object trajectories with visual representations during video generation to capture manipulation-relevant dynamics.

\section{Method}
\label{sec:method}

DIDO centers on interaction-centric dynamics: recovering them within a single denoising step, grounding them in manipulation-relevant entities, and efficiently exploiting them for action prediction. This perspective guides the token design and supervision of our one-step distillation, structuring the representation around the gripper, target object, their coupled evolution, and the target object's discriminative appearance. The training strategy and additional architectural details are provided in Section~\ref{sec:training} and the supplementary material.

\begin{figure*}[t]
\centering
\includegraphics[width=\textwidth]{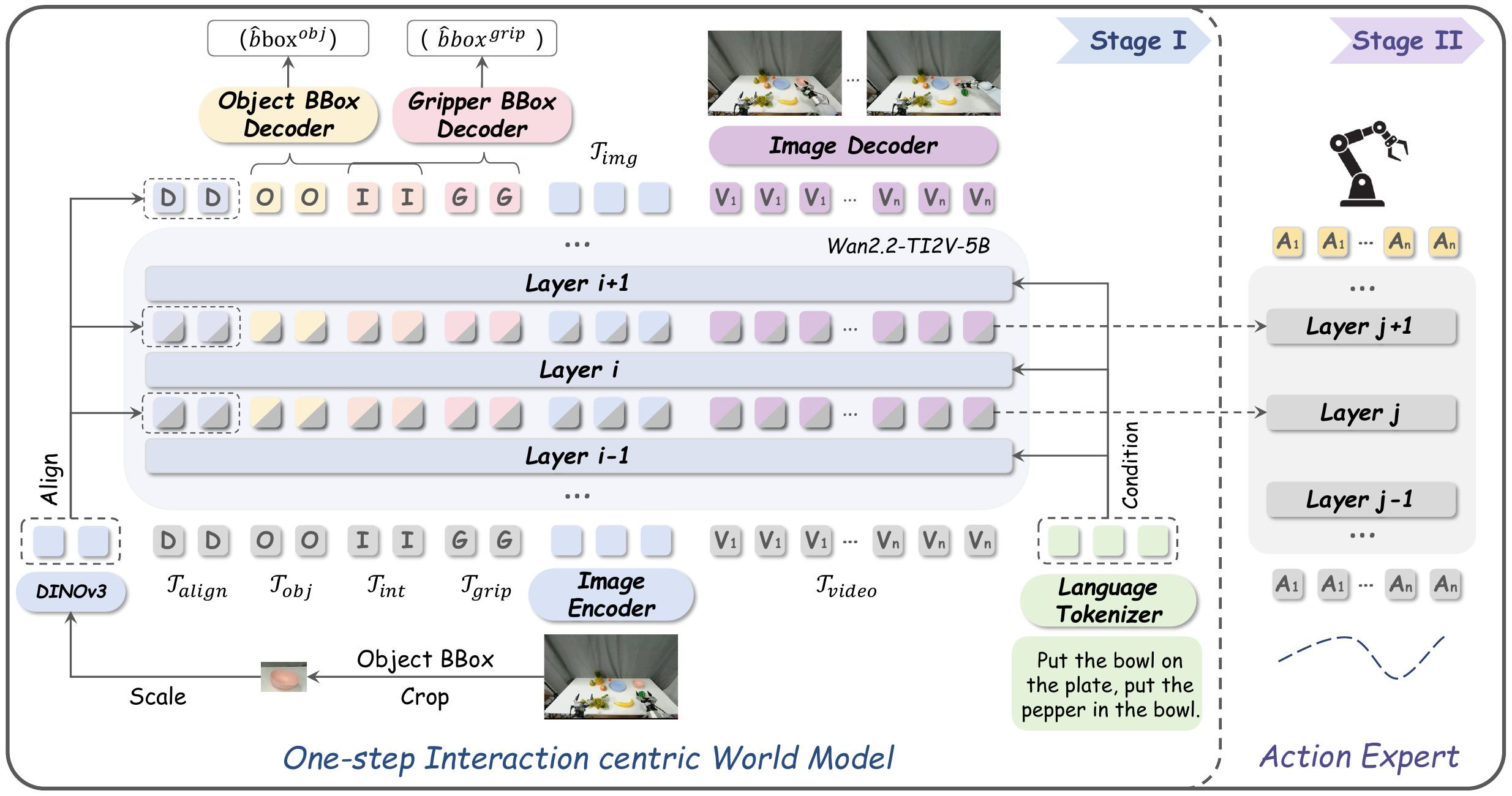}
\caption{
\textbf{Overview of DIDO.}
In Stage I, the one-step world model augments the video transformer with four interaction-centric token groups, including object $\mathcal{T}_{\mathrm{obj}}$, interaction $\mathcal{T}_{\mathrm{int}}$, gripper $\mathcal{T}_{\mathrm{grip}}$, and alignment $\mathcal{T}_{\mathrm{align}}$ tokens. The object and gripper branches predict future bounding-box trajectories $\hat{b}^{\mathrm{obj}}$ and $\hat{b}^{\mathrm{grip}}$ using the shared interaction tokens, while the alignment tokens are supervised by DINOv3 features extracted from the target object crop. Together with the image-conditioning and video representations $\mathcal{T}_{\mathrm{video}}$, these tokens evolve through the video transformer to capture interaction-centric future dynamics. In Stage II, an action expert is coupled to the world model through a mixture-of-transformers architecture and directly accesses its intermediate representations for action prediction without decoding and re-encoding future video.
}
\label{fig:pipeline}
\end{figure*}

\subsection{Interaction-Centric One-Step Distillation}
\label{sec:distill}

DIDO performs interaction-centric one-step distillation to recover manipulation-relevant interaction dynamics within a single denoising step. We build on Wan2.2-TI2V-5B~\citep{wan2025wan} by initializing the video model from its publicly released four-step distilled variant~\citep{li2025magicmotion} and adapting it to robot manipulation demonstrations as the teacher.

We distill the four-step teacher into a single-step generator using distribution matching distillation (DMD)~\citep{yin2024dmd2}. Let $p_{\mathrm{T},t}$ and $p_{\mathrm{\theta},t}$ denote the perturbed distributions induced by the four-step teacher and single-step generator at diffusion timestep $t$, respectively. DMD optimizes
\begin{equation}
\mathcal{L}_{\mathrm{DM}}
=
\mathbb{E}_{t}
\left[
D_{\mathrm{KL}}
\left(
p_{\mathrm{\theta},t}
\;\|\;
p_{\mathrm{T},t}
\right)
\right],
\label{eq:dmd}
\end{equation}
whose gradient is estimated using real and fake score networks. DMD trains the generator to match the output distribution of the multi-step teacher, enabling a single forward pass to approximate the teacher's converged future dynamics. However, generic distribution matching does not explicitly prioritize the coupled dynamics of the gripper and target object. To make the distillation interaction-centric, we introduce structured supervision for the manipulation-relevant entities, their interaction, and the target object's visual appearance.

Four interaction-centric token groups are defined as
\begin{equation}
\mathcal{T}_{\mathrm{inter}}
=
[
\mathcal{T}_{\mathrm{obj}};
\mathcal{T}_{\mathrm{int}};
\mathcal{T}_{\mathrm{grip}};
\mathcal{T}_{\mathrm{align}}
],
\end{equation}
representing the target object, gripper-object interaction, gripper, and object-centric visual alignment, respectively. These tokens are embedded directly into the video transformer and jointly optimized with the one-step distillation objective. They reflect our design principle that one-step distillation should prioritize manipulation-relevant information rather than allocate equal modeling capacity to all scene content: entity-specific tokens represent the gripper and target object, shared interaction tokens capture their coupled evolution, and alignment tokens preserve discriminative visual information about the manipulated object.

\paragraph{Trajectory Supervision.}
Bounding-box trajectories provide a direct and compact description of how the manipulation entities evolve over time. We therefore supervise the future spatial evolution of the gripper and target object over the action horizon. Let $H_a$ denote the action chunk size. The gripper branch $[\mathcal{T}_{\mathrm{grip}};\mathcal{T}_{\mathrm{int}}]$ predicts a sequence of $H_a$ future bounding boxes $\hat{b}^{\mathrm{grip}}=\{\hat{b}^{\mathrm{grip}}_h\}_{h=1}^{H_a}$, while the object branch $[\mathcal{T}_{\mathrm{obj}};\mathcal{T}_{\mathrm{int}}]$ predicts $\hat{b}^{\mathrm{obj}}=\{\hat{b}^{\mathrm{obj}}_h\}_{h=1}^{H_a}$. Crucially, $\mathcal{T}_{\mathrm{int}}$ is shared by both branches. It is therefore jointly supervised by the gripper and object trajectories, encouraging it to model their coupled spatial and temporal evolution.

For a predicted bounding-box sequence $\hat{b}=\{\hat{b}_h\}_{h=1}^{H_a}$ and its target $b=\{b_h\}_{h=1}^{H_a}$, we define
\begin{equation}
\mathcal{L}_{\mathrm{box}}(\hat{b},b)
=
\frac{1}{H_a}
\sum_{h=1}^{H_a}
\left[
\mathrm{smooth}_{L_1}(\hat{b}_h,b_h)
+
\lambda_{\mathrm{giou}}
\mathcal{L}_{\mathrm{giou}}(\hat{b}_h,b_h)
\right].
\label{eq:box}
\end{equation}
The interaction trajectory objective is
\begin{equation}
\mathcal{L}_{\mathrm{inter}}
=
\lambda_{\mathrm{obj}}
\mathcal{L}_{\mathrm{box}}
\left(
\hat{b}^{\mathrm{obj}},
b^{\mathrm{obj}}
\right)
+
\lambda_{\mathrm{grip}}
\mathcal{L}_{\mathrm{box}}
\left(
\hat{b}^{\mathrm{grip}},
b^{\mathrm{grip}}
\right).
\label{eq:interaction}
\end{equation}

\paragraph{Multi-Layer Visual Alignment.}
Manipulation also requires preserving discriminative visual information about the manipulated object. We therefore supervise $\mathcal{T}_{\mathrm{align}}$ with object-centric features extracted by a frozen DINOv3 encoder~\citep{simeoni2026dinov3}. Specifically, we crop the manipulated-object region using its bounding box, resize the crop to the DINOv3 input resolution, and encode it into a sequence of visual tokens. At selected video-transformer layers, the intermediate representations of $\mathcal{T}_{\mathrm{align}}$ are aligned with these frozen DINOv3 tokens.

Let $\mathcal{S}$ denote the transformer layers used for alignment, $\mathcal{T}_{\mathrm{align}}^{l}$ the alignment-token representations at layer $l$, and $F_{\mathrm{DINO}}$ the frozen DINOv3 tokens extracted from the object crop. We define
\begin{equation}
\mathcal{L}_{\mathrm{align}}
=
\frac{1}{|\mathcal{S}|}
\sum_{l\in\mathcal{S}}
\left\|
P_l
\left(
\mathcal{T}_{\mathrm{align}}^{l}
\right)
-
F_{\mathrm{DINO}}
\right\|_1,
\label{eq:visual_alignment}
\end{equation}
where $P_l$ projects the alignment-token representations into the DINOv3 feature space. The alignment objective complements trajectory supervision: trajectory supervision captures the spatial and temporal evolution of the manipulation entities, while visual alignment preserves the target object's discriminative appearance.

The required bounding-box trajectories and visual targets are automatically curated from the original demonstrations without manual labeling; details are provided in the supplementary material.

\subsection{Interaction-Aware Action Prediction}
\label{sec:action}

DIDO couples the single-step video model with an action expert through a mixture-of-transformers architecture, allowing the action expert to directly attend to the video model's internal representations without decoding and re-encoding future video. As illustrated in Figure~\ref{fig:pipeline}, the action expert accesses the complete latent stream of the video model, including image-conditioning, interaction-centric, and future-video representations. Only the future-video representations are selectively refined before being exposed to the action expert.

\paragraph{Dynamics-Based Token Refinement.}
\label{sec:refinement}
Future-video representations contain spatial redundancy, while our denoising analysis shows that interaction-related changes are localized. We therefore retain fine-grained tokens in high-change regions and pool the remaining regions to reduce the conditioning cost for action prediction. The concrete configuration varies with input resolution; we illustrate the procedure with RoboTwin.

Let $t_0$ denote the current frame and $t_k$ the $k$-th future frame, with $V_{t_0}$ and $V_{t_k}$ denoting their corresponding value features, $V_{t_0},V_{t_k}\in \mathbb{R}^{H \times W \times D}$. We partition the spatial feature map into a $4\times5$ grid of 20 regions, each containing a $2\times2$ set of fine-grained cells. For each cell $r=(h,w)$, we measure its predicted change as
 \begin{equation}
  s_{t_k}(r)
  =
  \frac{1}{D}
  \left\|
  \mathbf{V}_{t_k}(r,:)-\mathbf{V}_{t_0}(r,:)
  \right\|_1.
  \label{eq:motion_token_score}
  \end{equation}
For each region $R_m$, we average the scores of its four cells:
\begin{equation}
S_{t_k}(R_m)
=
\frac{1}{|R_m|}\sum_{r\in R_m}
s_{t_k}(r),
\qquad |R_m|=4.
\label{eq:region_motion}
\end{equation}
We rank the 20 regions by $S_{t_k}(R_m)$ and retain all fine-grained tokens in the top-3 regions. For each of the remaining 17 regions, $2\times2$ average pooling merges its four cells into a single token. The same spatial refinement is applied to the corresponding key and value features before being accessed by the action expert, while all other video-model latents remain unchanged.

\paragraph{Flow-Matching Action Generation.}
Conditioned on the video-model representations, the action expert generates an action chunk of length $H_a$ via flow matching. Let $a=\{a_h\}_{h=1}^{H_a}$ denote the ground-truth action chunk. We sample $\tau\sim U[0,1]$ and $\epsilon\sim\mathcal{N}(0,I)$ with the same shape as $a$, and construct $a_{\tau}=(1-\tau)\epsilon+\tau a$ with target velocity $u=a-\epsilon$. The action expert $v_{\theta}$ is optimized by
\begin{equation}
\mathcal{L}_{\mathrm{act}}
=
\mathbb{E}_{a,\tau,\epsilon}
\left[
\left\|
v_{\theta}
\left(
a_{\tau},\tau
\right)
-
u
\right\|_2^2
\right].
\label{eq:action}
\end{equation}

\subsection{Training Strategy}
\label{sec:training}

DIDO is trained in two stages after preparing the robot-domain four-step teacher described in Section~\ref{sec:distill}; detailed optimization settings are provided in the supplementary material.

\paragraph{Stage I: One-Step Interaction-Centric Dynamics Learning.}
The first stage jointly optimizes one-step distillation, interaction trajectory supervision, and visual alignment:
\begin{equation}
\mathcal{L}_{\mathrm{stage1}}
=
\mathcal{L}_{\mathrm{DM}}
+
\mathcal{L}_{\mathrm{inter}}
+
\lambda_{\mathrm{align}}
\mathcal{L}_{\mathrm{align}}.
\label{eq:stage1}
\end{equation}
Here, $\mathcal{L}_{\mathrm{DM}}$ transfers the teacher's converged future distribution, $\mathcal{L}_{\mathrm{inter}}$ grounds the prediction in the gripper and target-object trajectories, and $\mathcal{L}_{\mathrm{align}}$ preserves object-centric visual information.

\paragraph{Stage II: Joint Policy Learning.}
We then initialize the action expert and jointly optimize it with the single-step video model. To preserve predictive dynamics during policy learning, we retain the video-generation objective $\mathcal{L}_{\mathrm{video}}$~\citep{yuan2026fastwam} together with the interaction-centric objectives:
\begin{equation}
\mathcal{L}_{\mathrm{stage2}}
=
\lambda_{\mathrm{video}}
\mathcal{L}_{\mathrm{video}}
+
\lambda_{\mathrm{act}}
\mathcal{L}_{\mathrm{act}}
+
\mathcal{L}_{\mathrm{inter}}
+
\lambda_{\mathrm{align}}
\mathcal{L}_{\mathrm{align}}.
\label{eq:stage2}
\end{equation}
The video-model backbone, interaction-centric tokens, prediction heads, and action expert are jointly optimized, while the language encoder, VAE, and DINOv3 encoder remain frozen. DMD is used only in Stage I; the video generator remains single-step throughout Stage II and inference. 

\section{Experiments}

Our experiments are designed to address the following questions:
\noindent\textbf{RQ1}: How do denoising representations affect downstream manipulation control? (Section~\ref{sec:analysis})
\noindent\textbf{RQ2}: Does DIDO achieve strong performance and generalization across simulation and real-world manipulation? (Sections~\ref{sec:standard_performance} and~\ref{sec:robustness})
\noindent\textbf{RQ3}: What drives DIDO's performance gains, and how do its components affect accuracy and inference efficiency? (Section~\ref{sec:ablation})

\begin{table*}[t]
\centering
\small
\setlength{\tabcolsep}{0.8pt}
\renewcommand{\arraystretch}{1.0}
\caption{Performance comparison on LIBERO and RoboTwin. Results are reported as success rates (\%). PT indicates the use of pretraining. The best and second-best average results on each benchmark are shown in \textbf{bold} and \underline{underlined}, respectively.}
\label{tab:standard}
\begin{tabular*}{\textwidth}{@{\extracolsep{\fill}}lccccccccc@{}}
\toprule
& & \multicolumn{5}{c}{\textbf{LIBERO}} & \multicolumn{3}{c}{\textbf{RoboTwin}} \\
\cmidrule(lr){3-7} \cmidrule(l){8-10}
Method & Emb.PT & Spatial & Object & Goal & Long & Avg. & Clean & Rand. & Avg. \\
\midrule

\rowcolor{groupblue}
\multicolumn{10}{l}{\emph{Vision-Language-Action Models}} \\ \noalign{\vskip 0.7ex}
$\pi_{0}$~\citep{black2024pi0}
& \checkmark & 96.8 & 98.8 & 95.8 & 85.2 & 94.1 & 65.9 & 58.4 & 62.2 \\
$\pi_{0.5}$~\citep{intelligence2025pi05}
& \checkmark & 98.8 & 98.2 & 98.0 & 92.4 & 96.9 & 82.7 & 76.8 & 79.8 \\

\midrule
\rowcolor{groupblue}
\multicolumn{10}{l}{\emph{World Action Models}} \\ \noalign{\vskip 0.7ex}
LingBot-VA~\citep{li2026lingbotva}
& \checkmark & 98.5 & 99.6 & 97.2 & 98.5 & \underline{98.5} & 92.9 & 91.5 & \textbf{92.2} \\
Motus~\citep{bi2026motus}
& \checkmark & 96.8 & 99.8 & 96.6 & 97.6 & 97.7 & 88.7 & 87.0 & 87.8 \\
Faster-WAM~\citep{ma2026fasterwam}
& \xmark & 98.4 & 100.0 & 97.0 & 97.8 & \underline{98.5} & 89.7 & 88.6 & 89.2 \\
Flash-WAM~\citep{akbari2026flashwam}
& \xmark & 97.0 & 92.8 & 96.4 & 98.0 & 96.1 & 88.4 & 82.7 & 85.5 \\
Fast-WAM~\citep{yuan2026fastwam}
& \xmark & 98.2 & 100.0 & 97.0 & 95.2 & 97.6 & 91.9 & 91.8 & 91.8 \\
Fast-WAM-IDM~\citep{yuan2026fastwam}
& \xmark & 98.8 & 97.8 & 97.8 & 97.6 & 98.0 & 91.16 & 91.34 & 91.3 \\

\midrule
\textbf{DIDO (Ours)}
& \xmark & 99.8 & 99.6 & 98.6 & 98.0 & \textbf{99.0} & 91.7 & 92.0 & \underline{91.9} \\

\bottomrule
\end{tabular*}
\end{table*}

\subsection{Experimental Setup}

\paragraph{Benchmarks.}
We evaluate DIDO on two standard manipulation benchmarks, one simulation benchmark for generalization, and a real-robot setting. LIBERO~\citep{liu2023libero} contains four suites, Spatial, Object, Goal, and Long, covering spatial reasoning, object generalization, goal understanding, and long-horizon execution. RoboTwin~\citep{chen2025robotwin} is a dual-arm simulation benchmark with multi-camera observations and contact-rich bimanual tasks. We additionally use LIBERO-Plus~\citep{fei2025liberoplus} to evaluate robustness under controlled distribution shifts, and four real-world manipulation tasks to assess transfer beyond simulation. Simulation performance is reported as success rate, while real-world performance is measured by normalized task-progress scores. Detailed real-world tasks and scoring protocols are provided in the supplementary material.

\paragraph{Baselines.}
We compare DIDO with two groups of methods. The first includes VLA models that predict actions without explicit future visual modeling, including $\pi_{0}$~\citep{black2024pi0}, $\pi_{0.5}$~\citep{intelligence2025pi05}, and, on LIBERO-Plus, OpenVLA-OFT~\citep{kim2025openvlaoft} and X-VLA~\citep{zheng2026xvla}. For the primary WAM comparison, we use Fast-WAM~\citep{yuan2026fastwam} and Fast-WAM-IDM~\citep{yuan2026fastwam}, which provide closely related baselines with and without interaction-aware dynamics modeling. The tables additionally report other WAMs, including LingBot-VA~\citep{li2026lingbotva}, Motus~\citep{bi2026motus}, Faster-WAM~\citep{ma2026fasterwam}, Flash-WAM~\citep{akbari2026flashwam}, and ST-WAM~\citep{wang2026stwam}, to provide broader context.

\subsection{Analysis of Denoising Representations}
\label{sec:analysis}

\begin{figure}[!t]
    \centering
    \includegraphics[width=0.85\linewidth]{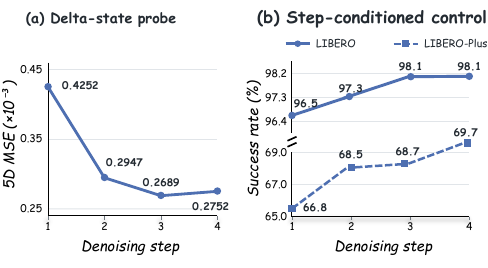}
    \caption{\textbf{Analysis of denoising representations.} Action evaluation and a frozen-feature delta-state probe show that later denoising representations generally provide stronger control signals and more predictable future state changes.}
    \label{fig:denoising_step_ablation}
\end{figure}

We first examine how denoising representations affect manipulation control. We use a four-step video model and an Action DiT expert without interaction-centric tokens or token refinement during training. At evaluation time, we vary only the denoising step from which the action expert receives its representation, while keeping the model parameters and all other settings fixed. In parallel, we freeze the four-step Wan video model before one-step distillation and train an identical probe for each step on a fixed split. Each probe maps hidden features from two temporal endpoints to a 5D future state delta (three position and two gripper coordinates), with MSE evaluated on 655 test pairs after training on 6,167 training pairs.

As shown in Figure~\ref{fig:denoising_step_ablation}, later representations improve downstream action success on both LIBERO and LIBERO-Plus, although the timing of the gain differs. On LIBERO, success increases at every step, from 97.7\% to 97.8\%, 98.1\%, and 98.4\%, for a total gain of 0.7 points. On LIBERO-Plus, success changes from 71.4\% to 71.3\% at Step~2, then rises to 71.7\% and 72.6\% at Steps~3 and~4; the 1.3-point recovery from Step~2 to~4 shows that the benefit emerges more clearly at later denoising stages under distribution shift.

The probe provides a complementary representation-level test. Its error decreases from $0.4252\times 10^{-3}$ to $0.2689\times 10^{-3}$ between Steps~1 and~3, a 36.8\% reduction, before a small rebound to $0.2752\times 10^{-3}$ at Step~4. Thus, the probe trend is not strictly monotonic, but every later step has lower error than Step~1, indicating more predictable future 5D state changes. The best probe result at Step~3 and the best control result at Step~4 need not coincide: they measure representation predictability and policy utility, respectively, while pointing to the same progressive emergence of control-relevant information.

The agreement between the action evaluation and the frozen-feature probe suggests that later denoising steps add information that is useful for control, rather than merely improving visual appearance. Directly truncating the process after the first step can therefore stop before part of the interaction-relevant predictive information has formed. This observation motivates DIDO's one-step world model, which is trained to recover these late-emerging dynamics without executing the full denoising trajectory.

\subsection{Simulation Experiments}
\label{sec:standard_performance}

\begin{table*}[t]
\centering
\caption{
LIBERO-Plus success rates (\%) under systematic perturbations.
For DIDO, Avg. is computed over all evaluation episodes.
Emb. PT. denotes embodied pretraining. The best and second-best average results are shown in \textbf{bold} and \underline{underlined}, respectively.
}
\label{tab:liberoplus}
\small
\setlength{\tabcolsep}{2.7pt}
\begin{tabular}{lccccccccc}
\toprule
Method & Emb. PT. & Camera & Robot & Lang. & Light & Backg. & Noise & Layout & Avg. \\
\midrule

\rowcolor{groupblue}
\multicolumn{10}{l}{\emph{Vision-Language-Action Models}} \\ \noalign{\vskip 0.7ex}
OpenVLA-OFT~\citep{kim2025openvlaoft}
    & \checkmark & 56.4 & 31.9 & 79.5 & 88.7 & 93.3 & 75.8 & 74.2 & 69.6 \\

$\pi_{0}$~\citep{black2024pi0}
    & \checkmark & 81.4 & 13.8 & 58.8 & 85.0 & 68.9 & 6.0 & 79.0 & 53.6 \\

X-VLA~\citep{zheng2026xvla}
    & \checkmark & 96.0 & 23.4 & 75.7 & 88.2 & 71.8 & 89.7 & 62.7 & 71.4 \\

\midrule
\rowcolor{groupblue}
\multicolumn{10}{l}{\emph{World Action Models}} \\ \noalign{\vskip 0.7ex}

Fast-WAM~\citep{yuan2026fastwam}
    & \xmark & 16.4 & 44.5 & 68.9 & 78.2 & 53.7 & 37.7 & 60.7 & 51.5 \\

Fast-WAM-IDM~\citep{yuan2026fastwam}
    & \xmark & 42.1 & 67.7 & 91.3 & 87.3 & 61.2 & 56.3 & 82.5 & 69.2 \\

ST-WAM~\citep{wang2026stwam}
    & \xmark & 55.4 & 60.1 & 79.3 & 93.0 & 74.2 & 79.5 & 74.3 & 72.8 \\

Faster-WAM~\citep{zhao2026fasterwam}
    & \xmark & 53.8 & 71.6 & 94.7 & 96.3 & 61.3 & 63.6 & 79.1 & \underline{73.6} \\

\midrule

\textbf{DIDO (Ours)}
    & \xmark & 43.5 & 75.2 & 94.5 & 97.0 & 72.4 & 73.8 & 83.0 & \textbf{76.3} \\

\bottomrule
\end{tabular}
\end{table*}
\begin{table*}[t]
\centering
\caption{\textbf{Component ablation and inference efficiency.}
Fast-WAM-IDM and one-step truncation provide the reference configurations (top); component ablations are reported below. Latency denotes end-to-end inference time; speedup is relative to Fast-WAM-IDM.}
\label{tab:ablation}
\begingroup
\small
\setlength{\tabcolsep}{4pt}
\renewcommand{\arraystretch}{1.12}
\begin{tabular*}{\textwidth}{@{\extracolsep{\fill}}lccc*{4}{c}@{}}
\toprule
& \multicolumn{3}{c}{Components} & \multicolumn{2}{c}{Success rate (\%) $\uparrow$} & \multicolumn{2}{c}{Efficiency} \\
\cmidrule(lr){2-4}\cmidrule(lr){5-6}\cmidrule(l){7-8}
Configuration & Distill. & Interact. & Refine. & LIBERO & LIBERO-Plus & Latency (ms) $\downarrow$ & Speedup $\uparrow$ \\
\midrule
\rowcolor{black!4}
Fast-WAM-IDM & \multicolumn{3}{c}{Full-model reference} & 98.0 & 69.2 & 871 & $1.00\times$ \\
\rowcolor{black!4}
One-step truncation & -- & -- & -- & 96.5 & 66.8 & 110 & $7.92\times$ \\
\midrule
Distillation only & $\checkmark$ & -- & -- & 97.9 & 71.4 & 110 & $7.92\times$ \\
Interaction only & -- & $\checkmark$ & -- & 98.4 & 71.3 & 117 & $7.44\times$ \\
Distill. + interaction & $\checkmark$ & $\checkmark$ & -- & 98.9 & 75.4 & 127 & $6.86\times$ \\
\addlinespace[2pt]
\rowcolor{groupblue}
\textbf{DIDO (full)} & $\checkmark$ & $\checkmark$ & $\checkmark$ & \textbf{99.0} & \textbf{76.3} & \textbf{108} & $\mathbf{8.06\times}$ \\
\bottomrule
\end{tabular*}
\endgroup
\end{table*}

\paragraph{LIBERO and RoboTwin.}
Table~\ref{tab:standard} evaluates single-arm and bimanual manipulation. On LIBERO, DIDO reaches 99.0\%, improving over Fast-WAM and Fast-WAM-IDM by 1.4 and 1.0 percentage points, respectively. On RoboTwin, DIDO achieves 91.9\%, which is 0.1 points above Fast-WAM and 0.6 points above Fast-WAM-IDM. These comparisons show that single-step future prediction remains competitive with closely related WAM baselines across both single-arm and bimanual settings. Co-video supervision ablations are reported in the supplementary material.

\paragraph{LIBERO-Plus.}
Table~\ref{tab:liberoplus} evaluates seven controlled distribution shifts. DIDO achieves an average success rate of 76.3\%, exceeding Fast-WAM and Fast-WAM-IDM by 24.8 and 7.1 percentage points, respectively. It ranks first under robot initialization, lighting, and object layout perturbations. The gains are not uniform: camera perturbations remain challenging, with DIDO scoring 43.5\% versus 42.1\% for Fast-WAM-IDM. Thus, the higher overall score reflects improvements across several perturbation types rather than uniformly superior robustness.

\begin{figure*}[t]
\centering
\includegraphics[width=\textwidth]{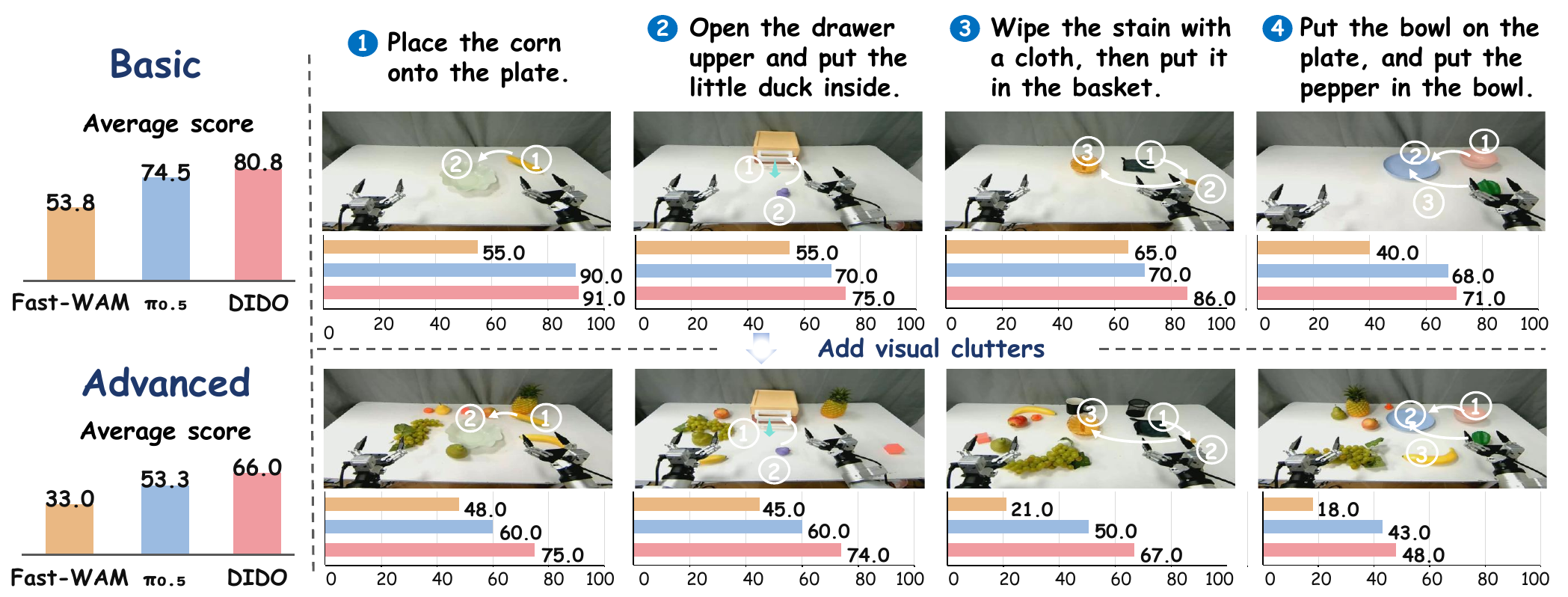}
\caption{
Real-world generalization on Galbot G1.
The basic protocol evaluates unseen target positions, while the advanced protocol additionally introduces visual clutter. Bars show normalized task scores for Fast-WAM, $\pi_{0.5}$, and DIDO, with the average across four tasks.
}
\label{fig:realworld}
\end{figure*}

\subsection{Real-World Experiments}
\label{sec:robustness}

We evaluate DIDO on Galbot~G1 across four manipulation tasks, with 150 demonstrations per task. The \emph{basic} protocol tests unseen target positions; the \emph{advanced} protocol additionally introduces task-irrelevant visual clutter. All methods use the same demonstrations, and policies are evaluated under both protocols without further adaptation; the full protocol is provided in the supplementary material.
As shown in Figure~\ref{fig:realworld}, DIDO achieves the highest scores on all four tasks under both protocols. Its average scores are 80.8 and 66.0 in the basic and advanced settings, compared with 74.5 and 53.3 for $\pi_{0.5}$, and 53.8 and 33.0 for Fast-WAM. The drop under added clutter is smaller for DIDO (14.8 points) than for $\pi_{0.5}$ (21.2) or Fast-WAM (20.8), supporting its robustness to combined position and appearance shifts.

\subsection{Ablation and Analysis}
\label{sec:ablation}

Table~\ref{tab:ablation} evaluates the contributions of DIDO's components against the one-step truncation baseline.

\paragraph{One-Step Distillation.}
Naively truncating the fine-tuned four-step video model to one denoising step achieves 96.5\% on LIBERO and 66.8\% on LIBERO-Plus. One-Step Dynamics Distillation improves performance to 97.9\% and 71.4\% at the same 110\,ms latency, supporting the recovery of control-relevant dynamics within a single step.

\paragraph{Interaction-Centric Tokens.}
Interaction-centric tokens with trajectory and visual supervision alone achieve 98.4\%/71.3\% on LIBERO/LIBERO-Plus, improving over truncation with only 7\,ms additional latency. Combining this supervision with distillation further achieves 98.9\%/75.4\%, exceeding either component alone by at least 4.0 points on LIBERO-Plus. These results support jointly supervising interaction dynamics during distillation: entity-level guidance helps the one-step model learn representations that are more useful for manipulation than either objective alone.

\begin{figure}[!t]
    \centering
    \includegraphics[width=0.50\linewidth]{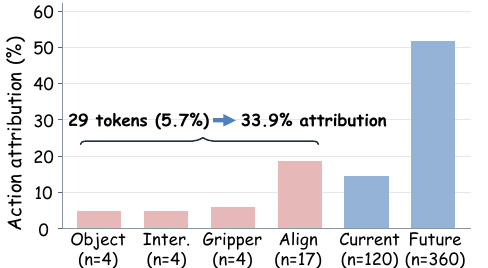}
    \caption{\textbf{Action attribution of interaction-centric tokens.} Only 29 interaction-centric tokens (5.74\% of the cache) receive 33.9\% of total action attribution.}
    \label{fig:token_attribution_simple}
\end{figure}

Additionally, we use action V-attribution to examine how the policy uses interaction-centric representations learned during distillation. As shown in Fig.~\ref{fig:token_attribution_simple}, these tokens occupy only 5.74\% of the cache but receive 33.9\% of total attribution, a $5.9\times$ enrichment relative to their token share. Together with the ablations, this suggests that interaction-centric supervision yields compact representations useful for control and actively used by the action expert. Metric definitions and detailed breakdowns are provided in Supplementary Appendix~D.3.

\paragraph{Dynamics-Based Token Refinement.}
Finally, Dynamics-Based Token Refinement reduces the conditioning context and lowers end-to-end latency from 127 to 108\,ms, while modestly improving success rates to 99.0\% and 76.3\%. Retaining dynamic regions while compressing near-static context thus preserves, and can improve, control performance, supporting the importance of manipulation-relevant dynamics for action prediction.

\paragraph{Inference Cost.}
The performance--efficiency panel of Figure~\ref{fig:trajectory} places DIDO in the high-success, low-latency region on LIBERO-Plus and RoboTwin. Table~\ref{tab:ablation} reports an end-to-end latency of 108\,ms for DIDO versus 871\,ms for Fast-WAM-IDM, corresponding to an approximately $8.1\times$ speedup. Against the one-step distillation baseline Flash-WAM, DIDO improves LIBERO/RoboTwin success by 2.9/6.4 points (Table~\ref{tab:standard}), with lower reported latency (108 vs.\ 322\,ms). These reported cross-method comparisons complement the controlled component ablations.

\section{Conclusion}

We introduced DIDO, an interaction-centric one-step distillation framework for World Action Models. Guided by the observation that manipulation-relevant interaction dynamics emerge late in denoising, DIDO jointly optimizes distribution matching and entity-level supervision to recover these dynamics in a single step. Beyond preserving performance under one-step compression, DIDO improves manipulation success over closely related WAM baselines while substantially reducing inference latency. These findings show that efficient predictive dynamics can retain manipulation-relevant interaction information through explicit interaction-centric supervision.

\bibliography{references}

\begin{thebibliography}{51}
\providecommand{\natexlab}[1]{#1}
\providecommand{\url}[1]{\texttt{#1}}
\expandafter\ifx\csname urlstyle\endcsname\relax
  \providecommand{\doi}[1]{doi: #1}\else
  \providecommand{\doi}{doi: \begingroup \urlstyle{rm}\Url}\fi

\bibitem[Akbari et~al.(2026)Akbari, Zhang, Akbari, Zhao, Chen, Chen, Zhang,
  Yuan, and Wang]{akbari2026flashwam}
Arman Akbari, Ci~Zhang, Arash Akbari, Lin Zhao, Yixiao Chen, Weiwei Chen, Xuan
  Zhang, Geng Yuan, and Yanzhi Wang.
\newblock Flash-wam: Modality-aware distillation for world action models.
\newblock \emph{arXiv preprint arXiv:2606.05254}, 2026.

\bibitem[Bai et~al.(2026{\natexlab{a}})Bai, Lyu, Zhou, Li, Wang, Xing, Zhao,
  Wang, Wang, Chi, Chen, and Zhang]{bai2026laravla}
Shuanghao Bai, Jing Lyu, Wanqi Zhou, Zhe Li, Dakai Wang, Lei Xing, Xiaoguang
  Zhao, Pengwei Wang, Zhongyuan Wang, Cheng Chi, Badong Chen, and Shanghang
  Zhang.
\newblock Latent reasoning vla: Latent thinking and prediction for
  vision-language-action models.
\newblock In \emph{Forty-third International Conference on Machine Learning},
  2026{\natexlab{a}}.

\bibitem[Bai et~al.(2026{\natexlab{b}})Bai, Song, Chen, Ji, Zhong, Yang, Zhao,
  Zhou, Li, Ding, et~al.]{bai2026embodied}
Shuanghao Bai, Wenxuan Song, Jiayi Chen, Yuheng Ji, Zhide Zhong, Jin Yang, Han
  Zhao, Wanqi Zhou, Zhe Li, Pengxiang Ding, et~al.
\newblock Embodied robot manipulation in the era of foundation models: Planning
  and learning perspectives.
\newblock \emph{IEEE Transactions on Robotics}, 2026{\natexlab{b}}.

\bibitem[Bi et~al.(2026)Bi, Tan, Xie, Wang, Huang, Liu, Zhao, Feng, Xiang,
  Rong, et~al.]{bi2026motus}
Hongzhe Bi, Hengkai Tan, Shenghao Xie, Zeyuan Wang, Shuhe Huang, Haitian Liu,
  Ruowen Zhao, Yao Feng, Chendong Xiang, Yinze Rong, et~al.
\newblock Motus: A unified latent action world model.
\newblock In \emph{Proceedings of the IEEE/CVF Conference on Computer Vision
  and Pattern Recognition}, pp.\  35101--35113, 2026.

\bibitem[Black et~al.(2024)Black, Nakamoto, Atreya, Walke, Finn, Kumar, and
  Levine]{black2024susie}
Kevin Black, Mitsuhiko Nakamoto, Pranav Atreya, Homer Walke, Chelsea Finn,
  Aviral Kumar, and Sergey Levine.
\newblock Zero-shot robotic manipulation with pre-trained image-editing
  diffusion models.
\newblock In \emph{International Conference on Learning Representations},
  volume 2024, pp.\  33431--33452, 2024.

\bibitem[Black et~al.(2025)Black, Brown, Driess, Esmail, Equi, Finn, Fusai,
  Groom, Hausman, Ichter, et~al.]{black2024pi0}
Kevin Black, Noah Brown, Danny Driess, Adnan Esmail, Michael Equi, Chelsea
  Finn, Niccolo Fusai, Lachy Groom, Karol Hausman, Brian Ichter, et~al.
\newblock $\pi_0$: A vision-language-action flow model for general robot
  control.
\newblock In \emph{Robotics: Science and Systems (RSS)}, 2025.

\bibitem[Brohan et~al.(2023)Brohan, Brown, Carbajal, Chebotar, Dabis, Finn,
  Gopalakrishnan, Hausman, Herzog, Hsu, et~al.]{brohan2023rt1}
Anthony Brohan, Noah Brown, Justice Carbajal, Yevgen Chebotar, Joseph Dabis,
  Chelsea Finn, Keerthana Gopalakrishnan, Karol Hausman, Alex Herzog, Jasmine
  Hsu, et~al.
\newblock Rt-1: Robotics transformer for real-world control at scale.
\newblock In \emph{Robotics: Science and Systems}, 2023.

\bibitem[Carion et~al.(2026)Carion, Gustafson, Hu, Debnath, Hu, Suris
  Coll-Vinent, Ryali, Alwala, Khedr, Huang, et~al.]{carion2026sam3}
Nicolas Carion, Laura Gustafson, Yuan-Ting Hu, Shoubhik Debnath, Ronghang Hu,
  Didac Suris Coll-Vinent, Chaitanya Ryali, Kalyan~Vasudev Alwala, Haitham
  Khedr, Andrew Huang, et~al.
\newblock Sam 3: Segment anything with concepts.
\newblock In \emph{International conference on learning representations},
  volume 2026, pp.\  138846--138923, 2026.

\bibitem[Cheang et~al.(2024)Cheang, Chen, Jing, Kong, Li, Li, Liu, Wu, Xu,
  Yang, et~al.]{cheang2024gr2}
Chi-Lam Cheang, Guangzeng Chen, Ya~Jing, Tao Kong, Hang Li, Yifeng Li, Yuxiao
  Liu, Hongtao Wu, Jiafeng Xu, Yichu Yang, et~al.
\newblock Gr-2: A generative video-language-action model with web-scale
  knowledge for robot manipulation.
\newblock \emph{arXiv preprint arXiv:2410.06158}, 2024.

\bibitem[Chen et~al.(2025)Chen, Chen, Chen, Cai, Liu, Li, Liang, Lin, Ge, Gu,
  et~al.]{chen2025robotwin}
Tianxing Chen, Zanxin Chen, Baijun Chen, Zijian Cai, Yibin Liu, Zixuan Li,
  Qiwei Liang, Xianliang Lin, Yiheng Ge, Zhenyu Gu, et~al.
\newblock Robotwin 2.0: A scalable data generator and benchmark with strong
  domain randomization for robust bimanual robotic manipulation.
\newblock \emph{arXiv preprint arXiv:2506.18088}, 2025.

\bibitem[Chi et~al.(2025)Chi, Ge, Liu, Zhou, Jia, He, Liu, Li, Han, Han,
  et~al.]{chi2025mind}
Xiaowei Chi, Kuangzhi Ge, Jiaming Liu, Siyuan Zhou, Peidong Jia, Zichen He,
  Yuzhen Liu, Tingguang Li, Lei Han, Sirui Han, et~al.
\newblock Mind: Learning a dual-system world model for real-time planning and
  implicit risk analysis.
\newblock \emph{arXiv preprint arXiv:2506.18897}, 2025.

\bibitem[Du et~al.(2023)Du, Yang, Dai, Dai, Nachum, Tenenbaum, Schuurmans, and
  Abbeel]{du2023unipi}
Yilun Du, Sherry Yang, Bo~Dai, Hanjun Dai, Ofir Nachum, Josh Tenenbaum, Dale
  Schuurmans, and Pieter Abbeel.
\newblock Learning universal policies via text-guided video generation.
\newblock In \emph{Advances in neural information processing systems},
  volume~36, pp.\  9156--9172, 2023.

\bibitem[Fei et~al.(2025)Fei, Wang, Shi, Dai, Cai, Qian, Ji, He, Zhang, Fei,
  Fu, Gong, and Qiu]{fei2025liberoplus}
Senyu Fei, Siyin Wang, Junhao Shi, Zihao Dai, Jikun Cai, Pengfang Qian, Li~Ji,
  Xinzhe He, Shiduo Zhang, Zhaoye Fei, Jinlan Fu, Jingjing Gong, and Xipeng
  Qiu.
\newblock Libero-plus: In-depth robustness analysis of vision-language-action
  models.
\newblock \emph{arXiv preprint arXiv:2510.13626}, 2025.

\bibitem[Hanyu et~al.(2026)Hanyu, Chung, Le, Nguyen, Ikebe, Gunderman, Nguyen,
  Vo, Kieu, Yamazaki, Rainwater, Nguyen, and Le]{hanyu2025slotvla}
Taisei Hanyu, Nhat Chung, Huy Le, Toan Nguyen, Yuki Ikebe, Anthony Gunderman,
  Duy Ho~Minh Nguyen, Khoa Vo, Tung Kieu, Kashu Yamazaki, Chase Rainwater, Anh
  Nguyen, and Ngan Le.
\newblock {SlotVLA}: Towards modeling of object-relation representations in
  robotic manipulation.
\newblock In \emph{Proceedings of the IEEE International Conference on Robotics
  and Automation (ICRA)}, 2026.

\bibitem[Hu et~al.(2025)Hu, Guo, Wang, Chen, Wang, Zhang, Sreenath, Lu, and
  Chen]{hu2025videopredictionpolicy}
Yucheng Hu, Yanjiang Guo, Pengchao Wang, Xiaoyu Chen, Yen-Jen Wang, Jianke
  Zhang, Koushil Sreenath, Chaochao Lu, and Jianyu Chen.
\newblock Video prediction policy: A generalist robot policy with predictive
  visual representations.
\newblock In \emph{Proceedings of the 42nd International Conference on Machine
  Learning}, pp.\  24328--24346, 2025.

\bibitem[Ji et~al.(2025)Ji, Tan, Shi, Hao, Zhang, Zhang, Wang, Zhao, Mu, An,
  Xue, Su, Lyu, Zheng, Liu, Wang, and Zhang]{ji2025robobrain}
Yuheng Ji, Huajie Tan, Jiayu Shi, Xiaoshuai Hao, Yuan Zhang, Hengyuan Zhang,
  Pengwei Wang, Mengdi Zhao, Yao Mu, Pengju An, Xinda Xue, Qinghang Su, Huaihai
  Lyu, Xiaolong Zheng, Jiaming Liu, Zhongyuan Wang, and Shanghang Zhang.
\newblock Robobrain: A unified brain model for robotic manipulation from
  abstract to concrete.
\newblock In \emph{Proceedings of the IEEE/CVF Conference on Computer Vision
  and Pattern Recognition (CVPR)}, pp.\  1724--1734, 2025.
\newblock \doi{10.1109/CVPR52734.2025.00168}.

\bibitem[Ji et~al.(2026)Ji, Liu, Tan, Huang, Huang, Xu, Chi, Zhao, Lyu, Co,
  Cao, Zhang, Li, Zhou, Wang, Wang, Zhang, and Zheng]{ji2026prmjudge}
Yuheng Ji, Yuyang Liu, Huajie Tan, Xuchuan Huang, Fanding Huang, Yijie Xu,
  Cheng Chi, Yuting Zhao, Huaihai Lyu, Peterson Co, Mingyu Cao, Qiongyu Zhang,
  Zhe Li, Enshen Zhou, Pengwei Wang, Zhongyuan Wang, Shanghang Zhang, and
  Xiaolong Zheng.
\newblock Prm-as-a-judge: A dense evaluation paradigm for fine-grained robotic
  auditing.
\newblock \emph{arXiv preprint arXiv:2603.21669}, 2026.

\bibitem[Jin et~al.(2026)Jin, Liao, Han, Li, and Chang]{jin2026structvla}
Minghao Jin, Mozheng Liao, Mingfei Han, Zhihui Li, and Xiaojun Chang.
\newblock Beyond dense futures: World models as structured planners for robotic
  manipulation.
\newblock In \emph{European Conference on Computer Vision}, 2026.

\bibitem[Kim et~al.(2025{\natexlab{a}})Kim, Finn, and Liang]{kim2025openvlaoft}
Moo~Jin Kim, Chelsea Finn, and Percy Liang.
\newblock Fine-tuning vision-language-action models: Optimizing speed and
  success.
\newblock In \emph{Robotics: Science and Systems}, 2025{\natexlab{a}}.

\bibitem[Kim et~al.(2025{\natexlab{b}})Kim, Pertsch, Karamcheti, Xiao,
  Balakrishna, Nair, Rafailov, Foster, Sanketi, Vuong, et~al.]{kim2025openvla}
Moo~Jin Kim, Karl Pertsch, Siddharth Karamcheti, Ted Xiao, Ashwin Balakrishna,
  Suraj Nair, Rafael Rafailov, Ethan~P Foster, Pannag~R Sanketi, Quan Vuong,
  et~al.
\newblock Openvla: An open-source vision-language-action model.
\newblock In \emph{Conference on Robot Learning}, pp.\  2679--2713. PMLR,
  2025{\natexlab{b}}.

\bibitem[Li et~al.(2026)Li, Zhang, Luo, Yang, Wang, Han, Yu, Gao, Xue, Zhu,
  et~al.]{li2026lingbotva}
Lin Li, Qihang Zhang, Yiming Luo, Shuai Yang, Ruilin Wang, Fei Han, Mingrui Yu,
  Zelin Gao, Nan Xue, Xing Zhu, et~al.
\newblock Causal world modeling for robot control.
\newblock In \emph{Robotics: Science and Systems}, 2026.

\bibitem[Li et~al.(2025{\natexlab{a}})Li, Wu, Xi, Li, Huang, Zhang, Chen, Wang,
  Zhu, Liu, and Huang]{li2025controlvla}
Puhao Li, Yingying Wu, Ziheng Xi, Wanlin Li, Yuzhe Huang, Zhiyuan Zhang,
  Yinghan Chen, Jianan Wang, Song-Chun Zhu, Tengyu Liu, and Siyuan Huang.
\newblock Controlvla: Few-shot object-centric adaptation for pre-trained
  vision-language-action models.
\newblock In \emph{Proceedings of The 9th Conference on Robot Learning}, pp.\
  1898--1913, 2025{\natexlab{a}}.

\bibitem[Li et~al.(2025{\natexlab{b}})Li, Xing, Wang, Zhang, Dai, and
  Wu]{li2025magicmotion}
Quanhao Li, Zhen Xing, Rui Wang, Hui Zhang, Qi~Dai, and Zuxuan Wu.
\newblock Magicmotion: Controllable video generation with dense-to-sparse
  trajectory guidance.
\newblock In \emph{2025 IEEE/CVF International Conference on Computer Vision
  (ICCV)}, pp.\  12112--12123. IEEE, 2025{\natexlab{b}}.

\bibitem[Li et~al.(2025{\natexlab{c}})Li, Xu, Zhang, Liu, Shen, Ponomarenko,
  Xu, Heng, Huang, Zhang, et~al.]{li2025crayonrobo}
Xiaoqi Li, Jingyun Xu, Mingxu Zhang, Jiaming Liu, Yan Shen, Iaroslav
  Ponomarenko, Jiahui Xu, Liang Heng, Siyuan Huang, Shanghang Zhang, et~al.
\newblock Object-centric prompt-driven vision-language-action model for robotic
  manipulation.
\newblock In \emph{2025 IEEE/CVF Conference on Computer Vision and Pattern
  Recognition (CVPR)}, pp.\  27638--27648. IEEE, 2025{\natexlab{c}}.

\bibitem[Liu et~al.(2023)Liu, Zhu, Gao, Feng, Liu, Zhu, and
  Stone]{liu2023libero}
Bo~Liu, Yifeng Zhu, Chongkai Gao, Yihao Feng, Qiang Liu, Yuke Zhu, and Peter
  Stone.
\newblock Libero: Benchmarking knowledge transfer for lifelong robot learning.
\newblock In \emph{Advances in Neural Information Processing Systems},
  volume~36, pp.\  44776--44791, 2023.

\bibitem[Liu et~al.(2024)Liu, Zeng, Ren, Li, Zhang, Yang, Jiang, Li, Yang, Su,
  et~al.]{liu2024groundingdino}
Shilong Liu, Zhaoyang Zeng, Tianhe Ren, Feng Li, Hao Zhang, Jie Yang, Qing
  Jiang, Chunyuan Li, Jianwei Yang, Hang Su, et~al.
\newblock Grounding dino: Marrying dino with grounded pre-training for open-set
  object detection.
\newblock In \emph{European conference on computer vision}, pp.\  38--55.
  Springer, 2024.

\bibitem[Liu et~al.(2026{\natexlab{a}})Liu, Sun, Li, Xie, Zhang, Chao, Dong,
  Chen, Zhang, and Ding]{liu2026oawam}
Yushan Liu, Peibo Sun, Shoujie Li, Yifan Xie, Lingfeng Zhang, Xintao Chao,
  Shiyuan Dong, Fang Chen, Xiao-Ping Zhang, and Wenbo Ding.
\newblock Oa-wam: Object-addressable world action model for robust robot
  manipulation.
\newblock \emph{arXiv preprint arXiv:2605.06481}, 2026{\natexlab{a}}.

\bibitem[Liu et~al.(2026{\natexlab{b}})Liu, Jie, Sun, Cao, Xu, Liu, and
  Chen]{liu2026sam3d}
Zonghe Liu, Shanyuan Jie, Xiaoquan Sun, Chen Cao, Zetian Xu, Zongsheng Liu, and
  Jiayu Chen.
\newblock Sam3d-guided object-centric representation alignment for
  vision-language-action models.
\newblock \emph{arXiv preprint arXiv:2607.25912}, 2026{\natexlab{b}}.

\bibitem[Lou et~al.(2026)Lou, Chi, Zhang, Qian, Li, Zhang, Lyu, Song, Fu, Xu,
  Wang, and Zhang]{lou2026maskworldmodel}
Yunfan Lou, Xiaowei Chi, Xiaojie Zhang, Zezhong Qian, Chengxuan Li, Rongyu
  Zhang, Yaoxu Lyu, Guoyu Song, Chuyao Fu, Haoxuan Xu, Pengwei Wang, and
  Shanghang Zhang.
\newblock Mask world model: Predicting what matters for robust robot policy
  learning.
\newblock In \emph{Proceedings of the International Conference on Machine
  Learning}, 2026.

\bibitem[Lyu et~al.(2026)Lyu, Liu, Zhang, Liao, Feng, Zhu, Shen, Chen, Zhang,
  Dong, et~al.]{lyu2026lda1b}
Jiangran Lyu, Kai Liu, Xuheng Zhang, Haoran Liao, Yusen Feng, Wenxuan Zhu,
  Tingrui Shen, Jiayi Chen, Jiazhao Zhang, Yifei Dong, et~al.
\newblock Lda-1b: Scaling latent dynamics action model via universal embodied
  data ingestion.
\newblock In \emph{Robotics: Science and Systems}, 2026.

\bibitem[Ma et~al.(2026)Ma, Yang, Zhang, Clemente, Hu, Cao, and
  Zhang]{ma2026fasterwam}
Liheng Ma, Rui~Heng Yang, Zhanguang Zhang, Mateo Clemente, Ziwen Hu, Tongtong
  Cao, and Yingxue Zhang.
\newblock Faster-wam: Do world action models need deep action modules?
\newblock \emph{arXiv preprint arXiv:2608.02365}, 2026.

\bibitem[{Physical Intelligence} et~al.(2025){Physical Intelligence}, Black,
  Brown, Darpinian, Dhabalia, Driess, Esmail, Equi, Finn, Fusai,
  et~al.]{intelligence2025pi05}
{Physical Intelligence}, Kevin Black, Noah Brown, James Darpinian, Karan
  Dhabalia, Danny Driess, Adnan Esmail, Michael Equi, Chelsea Finn, Niccolo
  Fusai, et~al.
\newblock {$\pi_{0.5}$}: A vision-language-action model with open-world
  generalization.
\newblock In \emph{Conference on Robot Learning}, pp.\  17--40. PMLR, 2025.

\bibitem[Sim{\'e}oni et~al.(2026)Sim{\'e}oni, Vo, Seitzer, Baldassarre, Oquab,
  Jose, Khalidov, Szafraniec, Yi, Ramamonjisoa, Massa, Haziza, Wehrstedt, Wang,
  Darcet, Moutakanni, Sentana, Roberts, Vedaldi, Tolan, Brandt, Couprie,
  Mairal, J{\'e}gou, Labatut, and Bojanowski]{simeoni2026dinov3}
Oriane Sim{\'e}oni, Huy~V. Vo, Maximilian Seitzer, Federico Baldassarre, Maxime
  Oquab, Cijo Jose, Vasil Khalidov, Marc Szafraniec, Seung~Eun Yi, Michael
  Ramamonjisoa, Francisco Massa, Daniel Haziza, Luca Wehrstedt, Jianyuan Wang,
  Timoth{\'e}e Darcet, Th{\'e}o Moutakanni, Leonel Sentana, Claire Roberts,
  Andrea Vedaldi, Jamie Tolan, John Brandt, Camille Couprie, Julien Mairal,
  Herv{\'e} J{\'e}gou, Patrick Labatut, and Piotr Bojanowski.
\newblock Dinov3.
\newblock \emph{Transactions on Machine Learning Research}, 2026.

\bibitem[Team et~al.(2026{\natexlab{a}})Team, Ye, Ma, Wang, Ni, Ye, Huang, Li,
  Zhao, Yan, et~al.]{team2026gigaworld05}
GigaWorld Team, Angen Ye, Angyuan Ma, Boyuan Wang, Chaojun Ni, Fangzheng Ye,
  Guan Huang, Guo Li, Guosheng Zhao, Haodong Yan, et~al.
\newblock Gigaworld-policy-0.5: A faster and stronger wam empowered by
  autoresearch.
\newblock \emph{arXiv preprint arXiv:2607.13960}, 2026{\natexlab{a}}.

\bibitem[Team et~al.(2026{\natexlab{b}})Team, Xiang, Bao, Liu, Tan, Bi, Li,
  Liu, Pang, Jing, et~al.]{team2026motubrain}
MotuBrain Team, Chendong Xiang, Fan Bao, Haitian Liu, Hengkai Tan, Hongzhe Bi,
  James Li, Jiabao Liu, Jingrui Pang, Kiro Jing, et~al.
\newblock Motubrain: An advanced world action model for robot control.
\newblock \emph{arXiv preprint arXiv:2604.27792}, 2026{\natexlab{b}}.

\bibitem[Tian et~al.(2025)Tian, Yang, Zeng, Wang, Lin, Dong, and
  Pang]{tian2025seer}
Yang Tian, Sizhe Yang, Jia Zeng, Ping Wang, Dahua Lin, Hao Dong, and Jiangmiao
  Pang.
\newblock Predictive inverse dynamics models are scalable learners for robotic
  manipulation.
\newblock In \emph{International Conference on Learning Representations},
  volume 2025, pp.\  92033--92052, 2025.

\bibitem[Wan et~al.(2025)Wan, Wang, Ai, Wen, Mao, Xie, Chen, Yu, Zhao, Yang,
  et~al.]{wan2025wan}
Team Wan, Ang Wang, Baole Ai, Bin Wen, Chaojie Mao, Chen-Wei Xie, Di~Chen,
  Feiwu Yu, Haiming Zhao, Jianxiao Yang, et~al.
\newblock Wan: Open and advanced large-scale video generative models.
\newblock \emph{arXiv preprint arXiv:2503.20314}, 2025.

\bibitem[Wang et~al.(2026{\natexlab{a}})Wang, Hu, Qian, Jiang, Wu, Yan, Jing,
  Hao, Wang, Niu, et~al.]{wang2026stwam}
Mingxin Wang, Bin Hu, Bin Qian, Kaitao Jiang, Haoning Wu, Feng Yan, Bowen Jing,
  Ruiyang Hao, Enyi Wang, Kangning Niu, et~al.
\newblock St-wam: Semantic-temporal world action model for robust manipulation
  under visual distribution shifts.
\newblock \emph{arXiv preprint arXiv:2607.28993}, 2026{\natexlab{a}}.

\bibitem[Wang et~al.(2026{\natexlab{b}})Wang, Ji, Cao, Shen, Xiao, Lyu, Xie,
  Liu, Tian, Long, et~al.]{wang2026orca}
Yihao Wang, Yuheng Ji, Mingyu Cao, Yanqing Shen, Runze Xiao, Huaihai Lyu,
  Senwei Xie, Euan Liu, Klara Tian, Tianfeng Long, et~al.
\newblock Orca: The world is in your mind.
\newblock \emph{arXiv preprint arXiv:2606.30534}, 2026{\natexlab{b}}.

\bibitem[Wu et~al.(2024)Wu, Jing, Cheang, Chen, Xu, Li, Liu, Li, and
  Kong]{wu2024gr1}
Hongtao Wu, Ya~Jing, Chilam Cheang, Guangzeng Chen, Jiafeng Xu, Xinghang Li,
  Minghuan Liu, Hang Li, and Tao Kong.
\newblock Unleashing large-scale video generative pre-training for visual robot
  manipulation.
\newblock In \emph{International Conference on Learning Representations},
  volume 2024, pp.\  10641--10662, 2024.

\bibitem[Yan et~al.(2026)Yan, Zhong, Zhu, He, Yuan, Song, Gong, Cai, Zhao, Yan,
  et~al.]{yan2026svam}
Haodong Yan, Zhide Zhong, Jiaguan Zhu, Junjie He, Weilin Yuan, Wenxuan Song,
  Xin Gong, Yingjie Cai, Guanyi Zhao, Xu~Yan, et~al.
\newblock S-vam: Shortcut video-action model by self-distilling geometric and
  semantic foresight.
\newblock In \emph{European Conference on Computer Vision}, 2026.

\bibitem[Ye et~al.(2026{\natexlab{a}})Ye, Wang, Ni, Huang, Zhao, Li, Li, Li,
  Lv, Liu, et~al.]{ye2026gigaworld}
Angen Ye, Boyuan Wang, Chaojun Ni, Guan Huang, Guosheng Zhao, Hao Li, Hengtao
  Li, Jie Li, Jindi Lv, Jingyu Liu, et~al.
\newblock Gigaworld-policy: An efficient action-centered world--action model.
\newblock \emph{arXiv preprint arXiv:2603.17240}, 2026{\natexlab{a}}.

\bibitem[Ye et~al.(2026{\natexlab{b}})Ye, Ge, Zheng, Gao, Yu, Kurian, Indupuru,
  Tan, Zhu, Xiang, et~al.]{ye2026dreamzero}
Seonghyeon Ye, Yunhao Ge, Kaiyuan Zheng, Shenyuan Gao, Sihyun Yu, George
  Kurian, Suneel Indupuru, You~Liang Tan, Chuning Zhu, Jiannan Xiang, et~al.
\newblock World action models are zero-shot policies.
\newblock \emph{arXiv preprint arXiv:2602.15922}, 2026{\natexlab{b}}.

\bibitem[Yin et~al.(2024)Yin, Gharbi, Park, Zhang, Shechtman, Durand, and
  Freeman]{yin2024dmd2}
Tianwei Yin, Micha{\"e}l Gharbi, Taesung Park, Richard Zhang, Eli Shechtman,
  Fredo Durand, and William~T Freeman.
\newblock Improved distribution matching distillation for fast image synthesis.
\newblock In \emph{Advances in neural information processing systems},
  volume~37, pp.\  47455--47487, 2024.

\bibitem[Yu et~al.(2026)Yu, Lin, Zhang, Zhang, Gu, Li, and Tan]{yu2026maskwam}
Hanyang Yu, Haitao Lin, Jingbo Zhang, Wenyao Zhang, Chenghao Gu, Heng Li, and
  Ping Tan.
\newblock Maskwam: Unifying mask prompting and prediction for world-action
  models.
\newblock \emph{arXiv preprint arXiv:2606.13515}, 2026.

\bibitem[Yuan et~al.(2026)Yuan, Dong, Liu, and Zhao]{yuan2026fastwam}
Tianyuan Yuan, Zibin Dong, Yicheng Liu, and Hang Zhao.
\newblock Fast-wam: Do world action models need test-time future imagination?
\newblock \emph{arXiv preprint arXiv:2603.16666}, 2026.

\bibitem[Zhang et~al.(2026)Zhang, Zhu, Su, Ma, Huang, Xu, and
  Wang]{zhang2026mecowam}
Jianjun Zhang, Jian Zhu, Taiyi Su, Chong Ma, Zitai Huang, Yi~Xu, and Hanli
  Wang.
\newblock Learning 4d geometric priors for inference-efficient world action
  models.
\newblock \emph{arXiv preprint arXiv:2607.05468}, 2026.

\bibitem[Zhao et~al.(2026)Zhao, Jiang, Shi, Liu, Huang, Su, Sui, and
  Wang]{zhao2026fasterwam}
Weiheng Zhao, Haoyi Jiang, Xin Shi, Liu Liu, Fan Huang, Zhizhong Su, Wei Sui,
  and Xinggang Wang.
\newblock Faster-wam: Efficient inference-time future conditioning for robust
  world action models.
\newblock \emph{arXiv preprint arXiv:2608.04404}, 2026.

\bibitem[Zheng et~al.(2026{\natexlab{a}})Zheng, Li, Wang, Liu, Kang, Feng,
  Zheng, Zou, Chen, Zeng, et~al.]{zheng2026xvla}
Jinliang Zheng, Jianxiong Li, Zhihao Wang, Dongxiu Liu, Xirui Kang, Yuchun
  Feng, Yinan Zheng, Jiayin Zou, Yilun Chen, Jia Zeng, et~al.
\newblock X-vla: Soft-prompted transformer as scalable cross-embodiment
  vision-language-action model.
\newblock In \emph{International Conference on Learning Representations},
  volume 2026, pp.\  60580--60606, 2026{\natexlab{a}}.

\bibitem[Zheng et~al.(2026{\natexlab{b}})Zheng, Lyu, Zhang, Chen, Yan, Deng,
  Shi, Zhao, Wang, Zhang, and Wang]{zheng2026emerging}
Yixin Zheng, Jiangran Lyu, Yifan Zhang, Jiayi Chen, Mi~Yan, Yuntian Deng,
  Xuesong Shi, Xiaoguang Zhao, Yizhou Wang, Zhizheng Zhang, and He~Wang.
\newblock Emerging extrinsic dexterity in cluttered scenes via dynamics-aware
  policy learning.
\newblock In \emph{Robotics: Science and Systems}, 2026{\natexlab{b}}.

\bibitem[Zitkovich et~al.(2023)Zitkovich, Yu, Xu, Xu, Xiao, Xia, Wu, Wohlhart,
  Welker, Wahid, et~al.]{zitkovich2023rt2}
Brianna Zitkovich, Tianhe Yu, Sichun Xu, Peng Xu, Ted Xiao, Fei Xia, Jialin Wu,
  Paul Wohlhart, Stefan Welker, Ayzaan Wahid, et~al.
\newblock Rt-2: Vision-language-action models transfer web knowledge to robotic
  control.
\newblock In \emph{Conference on Robot Learning}, pp.\  2165--2183. PMLR, 2023.

\end{thebibliography}
\bibliographystyle{iclr2027_conference}

\clearpage
\appendix
\section{Data Curation and Annotation}
\label{app:data}

Interaction-centric visual reasoning requires dense bounding-box trajectories for the gripper and manipulated object. Since the original datasets do not provide these annotations, we automatically curate them from the demonstrations without manual labeling.

\subsection{Automatic Interaction Annotation}

\paragraph{Entity Trajectory Annotation.}
For the manipulated object, we follow the automatic annotation pipeline of LaRA-VLA~\citep{bai2026laravla}. The target entity is first parsed from the language instruction and localized in a reference frame using an open-vocabulary detector. A video tracker then propagates the target through the sequence, producing dense frame-level bounding boxes.

For the gripper, LIBERO and RoboTwin provide its pose through simulator state, which is projected into image space to obtain the corresponding bounding boxes. For Galbot G1 in real-world experiments, where such state information is unavailable, the gripper is detected using Grounding DINO~\citep{liu2024groundingdino} and propagated through the sequence with SAM~3~\citep{carion2026sam3}. Tracking is performed in both temporal directions, and the higher-confidence prediction is retained at each frame. Missing intermediate boxes are filled by linear interpolation between valid detections.

\paragraph{Visual Alignment Targets.}
The manipulated-object annotations are also used to construct the targets for multi-layer visual alignment. For each frame, we crop the target-object region using its annotated bounding box, enlarge the crop by a fixed margin to retain local context, and resize it to the DINOv3 input resolution. A frozen DINOv3 encoder~\citep{simeoni2026dinov3} then extracts a sequence of object-centric visual tokens, which serve as fixed targets for the dedicated alignment tokens $\mathcal{T}_{\mathrm{align}}$.
Cropping before encoding keeps the supervision object-specific and reduces interference from static background content.

\paragraph{Annotation Quality.}
We audit the automatically generated bounding-box annotations at the trajectory level. On LIBERO, the target-object boxes remain correct throughout 95 of 100 sampled trajectories, while 5 exhibit tracking drift; the gripper boxes are correct in all sampled trajectories. On RoboTwin, the target-object boxes are correct in 181 of 200 sampled trajectories, with 9 detection failures and 10 tracking drifts, while the gripper boxes are correct in all cases. On Galbot~G1, all 40 sampled trajectories, with 10 per task, contain correct annotations for both the target object and gripper. These results indicate that the automatic pipeline provides reliable supervision for interaction-centric visual reasoning.

\section{Additional Model Details}
\label{app:model}

This section provides architectural details omitted from the main text.

\subsection{One-Step Distribution Matching Distillation}
\label{app:dmd_detail}
\paragraph{Video Model.}
DIDO builds on the pretrained Wan2.2-TI2V-5B video generation model \citep{wan2025wan} and is initialized from its publicly released four-step distilled checkpoint \citep{li2025magicmotion}. We further adapt this checkpoint to robot manipulation demonstrations, yielding a robot-domain four-step teacher used for distribution matching in Stage~I. The VAE decoder is used only to visualize predicted future frames and is not required during policy inference.

We distill the robot-domain four-step teacher into a one-step generator following the distribution-matching principle of DMD \citep{yin2024dmd2}. We denote the student generator by $G_{\theta}$, the frozen teacher score model by $f_{\mathrm{T}}$, and the online fake-score model by $f_{\mathrm{F}}$. All three networks are initialized from the same robot-domain four-step checkpoint. During distillation, $f_{\mathrm{T}}$ remains frozen, while $G_{\theta}$ and $f_{\mathrm{F}}$ are optimized alternately. Importantly, the frozen teacher is queried at a single perturbed student sample during each DMD update; it does not execute the complete four-step denoising trajectory.

Given conditioning information $c$, an observed first-frame latent $x_{\mathrm{obs}}$, and Gaussian noise $z_{\mathrm{future}}$ for the future latent frames, the student predicts the future-video latent in a single network evaluation:
\begin{equation}
    x_g = G_{\theta}
    \left(
        z_{\mathrm{future}},
        x_{\mathrm{obs}},
        c
    \right).
    \label{eq:onestep_generator}
\end{equation}
The observed first-frame latent is kept fixed, while the future latent frames are initialized from Gaussian noise.
\paragraph{Distribution Matching}

Let $p_{\theta}$ denote the distribution induced by the one-step student and $p_{\mathrm{T}}$ the target distribution represented by the frozen teacher. Rather than reproducing the teacher's multi-step sampling trajectory, we match their perturbed distributions:
\begin{equation}
    \mathcal{L}_{\mathrm{DM}}
    =
    \mathbb{E}_{t}
    \left[
        D_{\mathrm{KL}}
        \left(
            p_{\theta,t}
            \,\Vert\,
            p_{\mathrm{T},t}
        \right)
    \right],
    \label{eq:dmd_kl}
\end{equation}
where $p_{\theta,t}$ and $p_{\mathrm{T},t}$ denote the corresponding distributions at noise level $t$.

For a noisy distribution $p_t$, its score is defined as $s(x_t,t)=\nabla_{x_t}\log p_t(x_t)$. The gradient of Eq.~\eqref{eq:dmd_kl} with respect to the student parameters can be expressed using the difference between the fake and target scores:
\begin{equation}
    \nabla_{\theta}\mathcal{L}_{\mathrm{DM}}
    \propto
    \mathbb{E}
    \left[
        \left(
            s_{\mathrm{F}}(x_t,t,c)
            -
            s_{\mathrm{T}}(x_t,t,c)
        \right)
        \frac{\partial G_{\theta}}{\partial\theta}
    \right],
    \label{eq:dmd_score_grad}
\end{equation}
where timestep-dependent positive weighting factors are omitted for clarity. The score difference therefore provides a distribution-level direction for aligning the student with the teacher without explicitly evaluating probability densities.

\paragraph{Teacher-Aligned Score Estimation}

We sample a nominal timestep $\tau$ and map it to the corresponding scheduler noise level:
\begin{equation}
    \sigma_t = w(\tau),
    \label{eq:teacher_timesteps}
\end{equation}
where $w(\cdot)$ denotes the scheduler's timestep mapping. The mapping is applied by the Flow Matching scheduler and is not expanded here into a separate score-sampling schedule.

Given the student output $x_g$, we construct a noisy sample using the flow-matching forward process:
\begin{equation}
    x_t
    =
    (1-\sigma_t)x_g
    +
    \sigma_t\epsilon,
    \qquad
    \epsilon\sim\mathcal{N}(0,I).
    \label{eq:flow_noise}
\end{equation}
The same noisy sample is evaluated by the frozen teacher and the online fake-score model:
\begin{equation}
    \hat{x}^{\mathrm{T}}_0
    =
    f_{\mathrm{T}}(x_t,t,c),
    \qquad
    \hat{x}^{\mathrm{F}}_0
    =
    f_{\mathrm{F}}(x_t,t,c),
    \label{eq:real_fake_prediction}
\end{equation}
where the original Wan flow predictions are converted to clean-latent predictions before computing the distribution-matching signal. For the frozen teacher, the conditional and negative-prompt predictions are combined with classifier-free guidance; in our implementation the guidance scale is 6.0. The fake-score model uses a conditional prediction without an additional CFG branch.

For the corruption process in Eq.~\eqref{eq:flow_noise}, an $x_0$ predictor induces the score
\begin{equation}
    s(x_t,t)
    =
    \frac{
        (1-\sigma_t)\hat{x}_0-x_t
    }{
        \sigma_t^2
    }.
    \label{eq:x0_score}
\end{equation}
Therefore, for $0<\sigma_t<1$, the fake and teacher score difference becomes
\begin{equation}
    s_{\mathrm{F}}(x_t,t,c)
    -
    s_{\mathrm{T}}(x_t,t,c)
    =
    \frac{1-\sigma_t}{\sigma_t^2}
    \left(
        \hat{x}^{\mathrm{F}}_0
        -
        \hat{x}^{\mathrm{T}}_0
    \right).
    \label{eq:score_x0_relation}
\end{equation}
Thus, for interior noise levels with $0<\sigma_t<1$, the difference between their clean-latent predictions provides the same gradient direction as the theoretical score difference up to a timestep-dependent positive scaling factor. At the endpoint $\sigma_t=1$, this proportionality degenerates because the perturbed sample is pure noise; we retain the normalized clean-latent difference below as a practical endpoint surrogate.

Following the practical DMD formulation, we use the normalized gradient surrogate
\begin{equation}
    g_{\mathrm{DMD}}
    =
    \frac{
        \hat{x}^{\mathrm{F}}_0
        -
        \hat{x}^{\mathrm{T}}_0
    }{
        \operatorname{mean}
        \left|
            x_g-\hat{x}^{\mathrm{T}}_0
        \right|
    },
    \label{eq:practical_dmd}
\end{equation}
which reduces variation in gradient magnitude across noise levels. We inject $g_{\mathrm{DMD}}$ into $G_{\theta}$ through a stop-gradient surrogate objective such that the resulting gradient with respect to $x_g$ is proportional to $g_{\mathrm{DMD}}$. This aligns the one-step student with the distribution represented by the four-step teacher without requiring the student to reproduce its sampling trajectory.

\paragraph{Online Fake-Score Learning} 

As the student distribution evolves during training, the fake-score model $f_{\mathrm{F}}$ is updated online to track the current generator distribution. Given a student sample $x_g$, we independently sample a score timestep, perturb $x_g$ according to Eq.~\eqref{eq:flow_noise}, and optimize
\begin{equation}
    \mathcal{L}_{\mathrm{fake}}
    =
    \mathbb{E}
    \left[
        \left\|
            f_{\mathrm{F}}(x_t,t,c)
            -
            \operatorname{sg}(x_g)
        \right\|_2^2
    \right],
    \label{eq:fake_score_loss}
\end{equation}
where $\operatorname{sg}(\cdot)$ denotes the stop-gradient operator. Following DMD2, $f_{\mathrm{F}}$ is updated more frequently than $G_{\theta}$ so that it can track the evolving student distribution and provide a stable distribution-matching signal.

In our object-aware implementation, the generator additionally receives auxiliary object, gripper, and DINO supervision. These auxiliary losses are applied only to the generator and are not used to train the frozen teacher or the fake-score model. Real robot videos provide the observed first-frame condition and the object-aware supervision targets; the DMD configuration does not apply a full future-video latent reconstruction loss.

During training, the frozen four-step teacher provides the target score estimate, while $f_{\mathrm{F}}$ estimates the score of the current one-step student distribution. At inference time, both score models are discarded and only $G_{\theta}$ is retained, reducing four-step future prediction to a single network evaluation.

\subsection{Other Model Details}

\paragraph{Interaction-Centric Tokens.}
DIDO uses 29 interaction-centric tokens divided into four groups: 4 object tokens $\mathcal{T}_{\mathrm{obj}}$, 4 interaction tokens $\mathcal{T}_{\mathrm{int}}$, 4 gripper tokens $\mathcal{T}_{\mathrm{grip}}$, and 17 alignment tokens $\mathcal{T}_{\mathrm{align}}$. These tokens are inserted into the video transformer and evolve jointly with its latent representations.

For trajectory prediction, $[\mathcal{T}_{\mathrm{obj}};\mathcal{T}_{\mathrm{int}}]$ and $[\mathcal{T}_{\mathrm{grip}};\mathcal{T}_{\mathrm{int}}]$ predict future object and gripper boxes, respectively. Each branch uses a two-layer MLP with GELU activations and hidden width equal to the video-transformer width, predicting normalized bounding-box coordinates for $H_a=32$ future steps. For visual alignment, the representations $\mathcal{T}_{\mathrm{align}}^{l}$ at selected transformer layers are mapped by lightweight projections $P_l$ to the DINOv3 feature space and supervised by Equation~\ref{eq:visual_alignment}. DINOv3 is used only to construct training targets and is removed at inference.

\begin{figure*}[t]
\centering
\includegraphics[width=\textwidth]{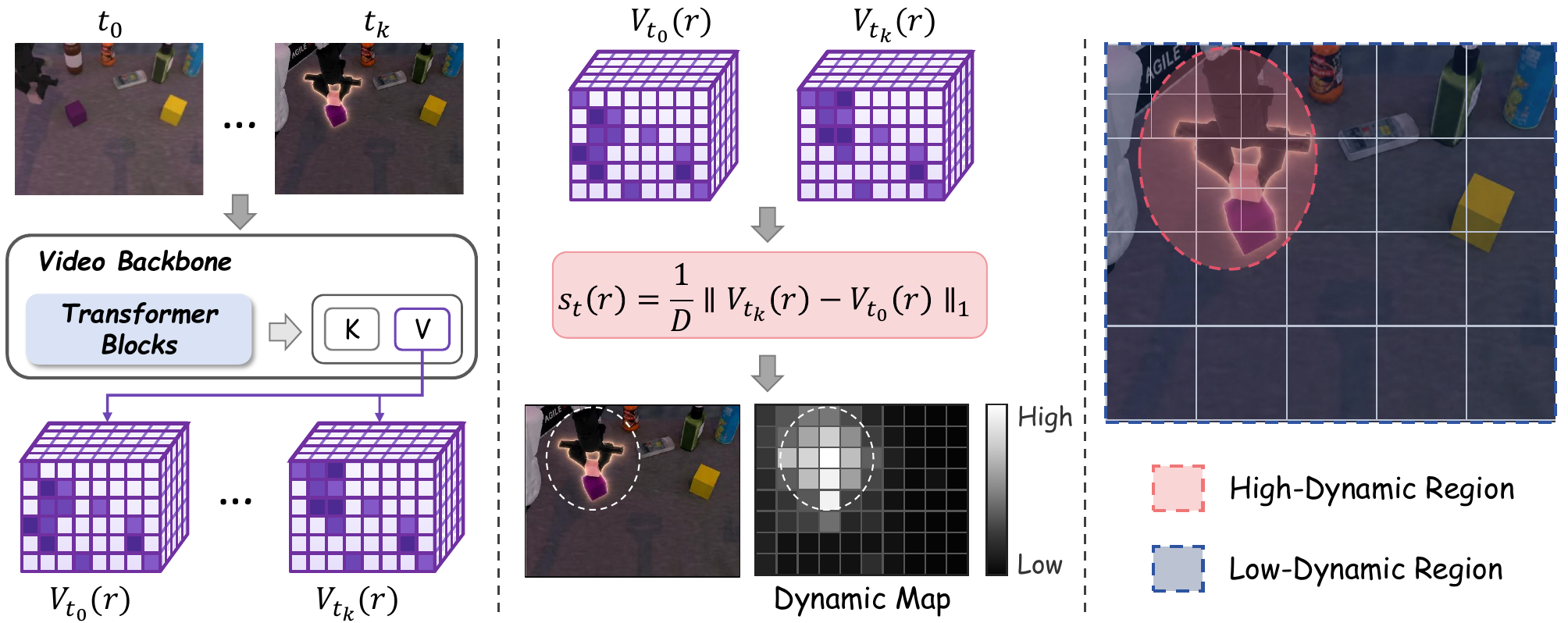}
\caption{
\textbf{Dynamics-based token refinement.}
Given the current timestamp $t_0$ and a future timestamp $t_k$, we obtain their corresponding value features $V_{t_0}$ and $V_{t_k}$ from the video backbone and compute cell-wise differences to construct a dynamic map. The spatial feature map is partitioned into a grid, with each region containing $2\times2$ fine-grained cells. We rank the regions by the averaged dynamics of their four cells, retain the top-3 high-dynamic regions at full token resolution, and apply $2\times2$ average pooling to the remaining regions. This preserves fine-grained interaction-relevant dynamics while substantially reducing redundant future-video tokens.
}
\label{fig:token_pruning}
\end{figure*}

\begin{figure}[t]
\centering
\includegraphics[width=0.50\linewidth]{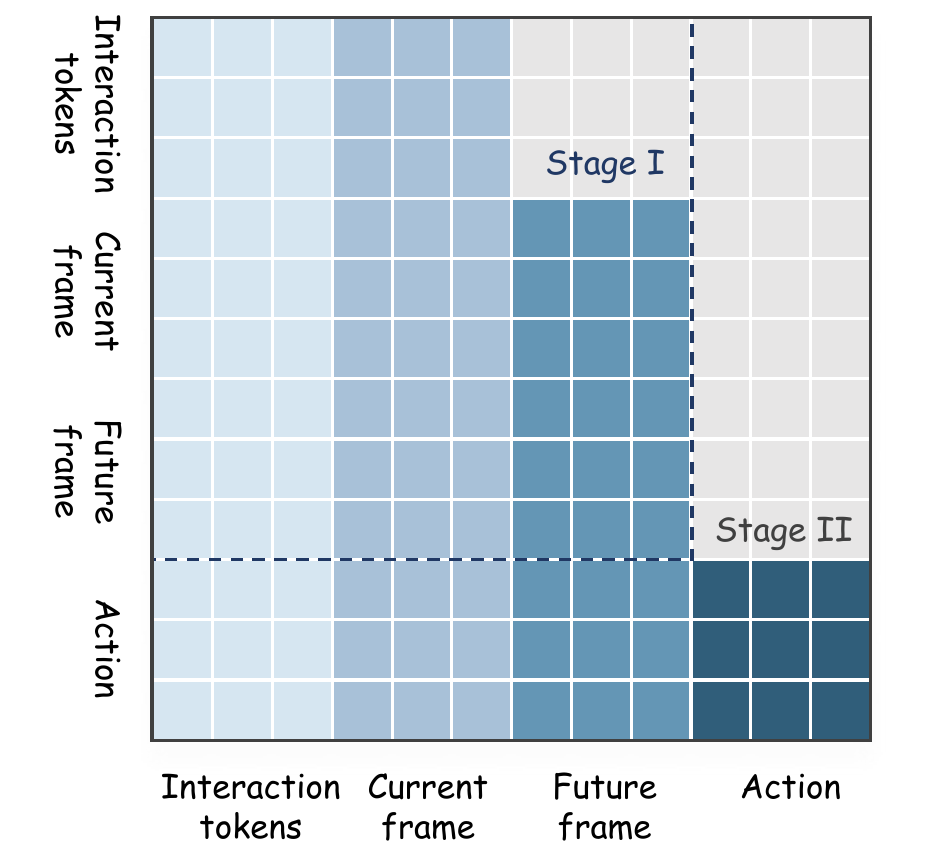}
\caption{Shared attention between the world model and action expert.}
\label{fig:attention}
\end{figure}

\paragraph{Action Expert.}
The action expert has the same depth as the video model and one quarter of its hidden width. It is coupled to the video backbone through the mixture-of-transformers architecture described in Section~\ref{sec:action}. Its output projection matches the action dimension of each embodiment, which is 7 for LIBERO, 14 for dual-arm RoboTwin, and 14 for Galbot~G1. The action expert predicts a chunk of $H_a=32$ actions per forward pass. During inference, the flow field is integrated with the default action-integration budget used for the reported results, and the first 10 actions on LIBERO and 24 actions on RoboTwin are executed before replanning.

\paragraph{Shared Attention.}
The video model and action expert are coupled through shared attention at corresponding transformer layers, as shown in Figure~\ref{fig:attention}. In Stage I, interaction-centric, current-frame, and future-frame tokens attend within the world-model stream. In Stage II, action tokens are introduced and can attend to all world-model representations and action tokens, whereas world-model tokens are prevented from attending to action tokens. This asymmetric attention gives the action expert direct access to intermediate world-model representations while preserving the predictive pathway learned in Stage~I.

\section{Additional Experimental Setup}
\label{app:exp_setup}

This section summarizes the training data (Section~\ref{app:training_data}), optimization details and hyperparameters (Section~\ref{app:impl}), and evaluation protocols (Section~\ref{app:protocol}).

\subsection{Denoising Analysis and Frozen-Feature Delta-State Probe}
\label{app:analysis_details}

\paragraph{Action evaluation.}
For the action evaluation in Section~\ref{sec:analysis}, we use the same four-step video model and Action DiT expert throughout the comparison. Interaction-centric tokens and token refinement are disabled, and only the denoising representation supplied to the action expert is changed at evaluation time. All other model parameters, evaluation episodes, and action-generation settings are fixed across Steps 1--4.

\paragraph{Frozen-feature delta-state probe.}
For the representation analysis, we freeze the four-step Wan video model before one-step distillation and train an identically configured probe separately for each denoising step. Each probe predicts the future 5D robot-state change from the corresponding hidden features and is evaluated with mean squared error. The complete fixed split contains 6,167 training pairs and 655 test pairs, and the same split, optimization budget, and probe architecture are used for every denoising step. All probes use AdamW with learning rate $1\times10^{-4}$, weight decay $1\times10^{-4}$, batch size 64, five training epochs, and seed 1729; no validation split or validation-based checkpoint selection is used. Thus, the comparison measures the predictive content of the representations rather than differences in probe capacity, data, or fitting procedure.

\paragraph{Probe architecture and target.}
Each input pair contains cached DiT hidden features from two temporal endpoints, with 30 transformer blocks, 98 spatial tokens, and 3,072 channels per endpoint. Learned channel compression reduces the channel dimension as $3072\!\rightarrow\!1536\!\rightarrow\!768\!\rightarrow\!384$ while preserving all tokens. Endpoint, block, camera, and spatial embeddings encode token identity. Eight learnable queries aggregate the compressed tokens using eight-head cross-attention, followed by a residual feed-forward block and an MLP prediction head. The five-dimensional target contains three position-delta coordinates and two gripper-delta coordinates. Targets are standardized with training-set statistics, and the reported MSE is averaged over all five coordinates.

\subsection{Training Data}
\label{app:training_data}

Beyond the pretrained Wan2.2 initialization, we use only the robot demonstrations provided by the corresponding evaluation settings, without additional robot video or cross-embodiment data. For LIBERO, both Stage~I and Stage~II are jointly trained on demonstrations from all four suites, and a single model is evaluated across all suites. RoboTwin follows the same setting, with both stages trained on the full training set and a single model evaluated across all tasks. For the real-world experiments, Stage~I uses all 600 demonstrations across four tasks, with 150 demonstrations per task, while Stage~II trains a separate policy for each task using its corresponding 150 demonstrations. LIBERO, RoboTwin, and the real-world experiments are trained independently and share no policy checkpoints.

\subsection{Implementation Details}
\label{app:impl}

Training consists of three optimization phases: teacher preparation, Stage~I, and Stage~II. All phases use AdamW with weight decay $0.01$ and bf16 mixed precision. Detailed hyperparameters are summarized in Table~\ref{tab:hparams}.

\begin{table}[t]
\centering
\footnotesize
\caption{Hyperparameter settings for teacher preparation and the two training stages of DIDO on LIBERO, RoboTwin, and real-world experiments.}
\label{tab:hparams}
\resizebox{\columnwidth}{!}{%
\begin{tabular}{lccc}
\toprule
\textbf{Hyperparameter} & \textbf{LIBERO} & \textbf{RoboTwin} & \textbf{Real World} \\
\midrule

\multicolumn{4}{l}{\textbf{Shared settings}} \\
\addlinespace[1pt]
Optimizer & AdamW & AdamW & AdamW \\
Weight decay & 0.01 & 0.01 & 0.01 \\
Precision & bf16 & bf16 & bf16 \\

\midrule
\multicolumn{4}{l}{\textbf{Teacher preparation}} \\
\addlinespace[1pt]
Learning rate & $3{\times}10^{-6}$ & $3{\times}10^{-6}$ & $3{\times}10^{-6}$ \\
Betas & $(0.0,0.999)$ & $(0.0,0.999)$ & $(0.0,0.999)$ \\
Schedule & 500 warm-up, const & 500 warm-up, const & 100 warm-up, const \\
Training steps & 3K & 4K & 1K \\
Global batch size & 384 & 2304 & 384 \\
$\lambda_{\mathrm{obj}}$ / $\lambda_{\mathrm{grip}}$ / $\lambda_{\mathrm{giou}}$
& 0.05 / 0.05 / 0.5
& 0.05 / 0.05 / 0.5
& 0.05 / 0.05 / 0.5 \\
$\lambda_{\mathrm{align}}$
& 0.02 & 0.02 & 0.02 \\

\midrule
\multicolumn{4}{l}{\textbf{Stage~\stageI: one-step dynamics learning}} \\
\addlinespace[1pt]
Generator LR & $1{\times}10^{-6}$ & $1{\times}10^{-6}$ & $1{\times}10^{-6}$ \\
Fake score LR & $1{\times}10^{-7}$ & $1{\times}10^{-7}$ & $1{\times}10^{-7}$ \\
Betas & $(0.0,0.999)$ & $(0.0,0.999)$ & $(0.0,0.999)$ \\
Schedule & 500 warm-up, const & 500 warm-up, const & 100 warm-up, const \\
Generator update interval & 5 iters & 5 iters & 5 iters \\
Training steps & 3K & 3K & 1K \\
Global batch size & 128 & 1536 & 128 \\
$\lambda_{\mathrm{obj}}$ / $\lambda_{\mathrm{grip}}$ / $\lambda_{\mathrm{giou}}$
& 0.03 / 0.03 / 0.5
& 0.045 / 0.045 / 0.5
& 0.03 / 0.03 / 0.5 \\
$\lambda_{\mathrm{align}}$
& 0.01 & 0.01 & 0.01 \\

\midrule
\multicolumn{4}{l}{\textbf{Stage~\stageII: joint policy learning}} \\
\addlinespace[1pt]
Learning rate & $1{\times}10^{-4}$ & $1{\times}10^{-4}$ & $1{\times}10^{-4}$ \\
Betas & $(0.9,0.95)$ & $(0.9,0.95)$ & $(0.9,0.95)$ \\
Schedule & 5\% warm-up, cosine & 5\% warm-up, cosine & 5\% warm-up, cosine \\
Training steps & 14{,}480 & 90K & 6K \\
Global batch size & 384 & 384 & 96 \\
$\lambda_{\mathrm{obj}}$ / $\lambda_{\mathrm{grip}}$ / $\lambda_{\mathrm{giou}}$
& 0.03 / 0.03 / 1.0
& 0.03 / 0.03 / 1.0
& 0.03 / 0.03 / 1.0 \\
$\lambda_{\mathrm{align}}$
& 0.02 & 0.02 & 0.02 \\
$\lambda_{\mathrm{video}}$ / $\lambda_{\mathrm{act}}$
& 0.5 / 1.0 & 0.5 / 1.0 & 0.5 / 1.0 \\

\bottomrule
\end{tabular}%
}
\vskip 0.1in
\end{table}

\paragraph{Teacher Preparation.}
Before the two-stage training of DIDO, we adapt the publicly released four-step Wan2.2-TI2V-5B checkpoint to robot manipulation videos. Interaction-centric supervision is enabled during this adaptation, with the corresponding warm-up schedules specified in Table~\ref{tab:hparams}. The resulting robot-domain four-step model serves as the teacher for Stage~I 

\paragraph{Stage I Optimization.}
We initialize the one-step generator and both score networks from the robot-domain teacher. The teacher score network is frozen; the fake-score network is updated every iteration, and the generator every fifth iteration. We use classifier-free guidance with scale 6 for the teacher score and no guidance for the fake-score network. The DMD objective and normalized gradient are described in Appendix~\ref{app:dmd_detail}. Stage-I learning rates, update intervals, and training steps are listed in Table~\ref{tab:hparams}.

\paragraph{Stage II Optimization.}
In Stage~II, the text encoder, VAE, and DINOv3 encoder remain frozen, while the video-transformer backbone, interaction-centric tokens, their prediction and projection heads, and action expert are jointly optimized. The DMD objective is removed and replaced by the video-generation objective following Fast-WAM~\citep{yuan2026fastwam}, retaining future prediction as a training signal during policy learning.

Let $z$ denote the ground-truth future video latents encoded by the frozen VAE, and let $\mathcal{C}_{\mathrm{vid}}$ denote the video-model conditioning. We follow the general flow matching objective, with 
$\tau=0$ enforced as our one-step generator does not require intermediate timesteps. For $\epsilon_z\sim\mathcal{N}(0,I)$, we construct $z_{\tau}=(1-\tau)\epsilon_z+\tau z$ with target velocity $u_z=z-\epsilon_z$. The video-generation objective is
\begin{equation}
\mathcal{L}_{\mathrm{video}}
=
\mathbb{E}_{z,\tau,\epsilon_z}
\left[
\left\|
v_{\phi}
\left(
z_{\tau},\tau
\mid
\mathcal{C}_{\mathrm{vid}}
\right)
-
u_z
\right\|_2^2
\right],
\label{eq:video}
\end{equation}
where $v_{\phi}$ denotes the video model. Video generation, trajectory supervision, visual alignment, and action prediction are jointly optimized during Stage~II.

\paragraph{Loss Weights and Hyperparameters.}
For each predicted bounding box, the smooth-$L_1$ term uses $\beta=1.0$, while the GIoU term is weighted by $\lambda_{\mathrm{giou}}$. The resulting object and gripper box losses are further scaled by $\lambda_{\mathrm{obj}}$ and $\lambda_{\mathrm{grip}}$, respectively, to form $\mathcal{L}_{\mathrm{inter}}$. The multi-layer visual alignment loss $\mathcal{L}_{\mathrm{align}}$ is weighted by $\lambda_{\mathrm{align}}$. In Stage~II, the video-generation loss $\mathcal{L}_{\mathrm{video}}$ and action loss $\mathcal{L}_{\mathrm{act}}$ are additionally weighted by $\lambda_{\mathrm{video}}$ and $\lambda_{\mathrm{act}}$, respectively. All loss weights and optimization hyperparameters are summarized in Table~\ref{tab:hparams}.

\paragraph{Compute.}
All experiments are trained on NVIDIA H100 GPUs with the global batch sizes reported in Table~\ref{tab:hparams}. For the simulation benchmarks, LIBERO requires approximately 15 hours for teacher preparation, 5 hours for Stage~I, and 7 hours for Stage~II. RoboTwin requires approximately 25 hours, 12 hours, and 100 hours for the three phases, respectively, primarily due to its larger demonstration set and optimization budget.

\subsection{Evaluation Protocol}
\label{app:protocol}

\begin{figure*}[t]
\centering
\includegraphics[width=\textwidth]{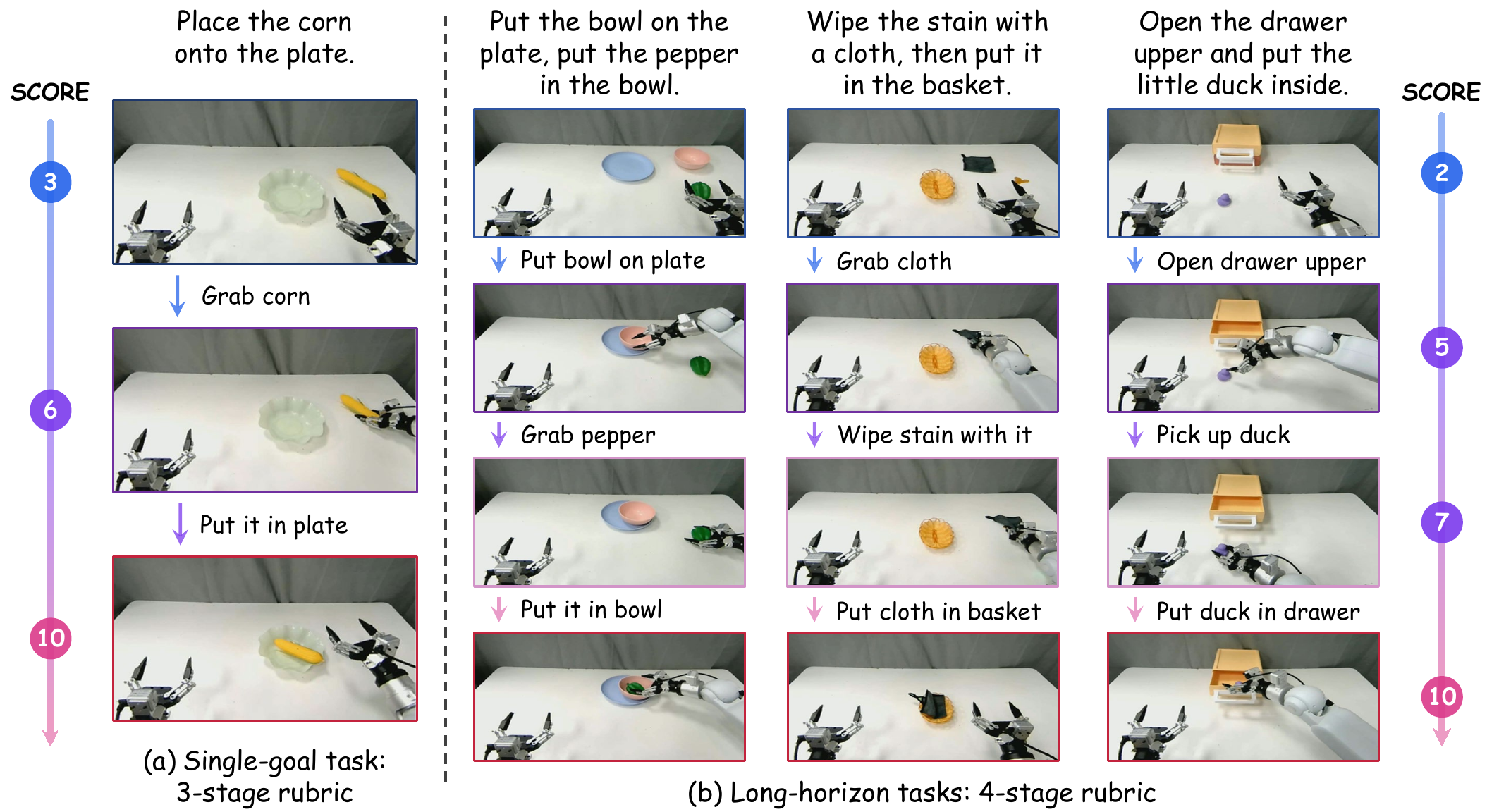}
\caption{
Staged scoring for real-world tasks.
}
\label{fig:rubric}
\end{figure*}

\paragraph{Simulation Evaluation.}
LIBERO, RoboTwin, and LIBERO-Plus follow their standard evaluation protocols. Performance is reported as task success rate averaged over evaluation episodes.

\paragraph{Real-World Evaluation.}
As shown in Figure~\ref{fig:rubric}, we evaluate DIDO on four real-world manipulation tasks: (i) Corn on Plate, placing the corn onto the plate; (ii) Bowl on Plate and Pepper in Bowl, placing the bowl on the plate and then putting the pepper in the bowl; (iii) Wipe Stain and Store Cloth, wiping the stain with a cloth and then putting the cloth in the basket; and (iv) Open Drawer and Place Duck, opening the upper drawer and putting the little duck inside. The first is a single-goal task evaluated with a three-stage rubric, while the remaining three are long-horizon tasks evaluated with four-stage rubrics.

Binary success does not distinguish between failures occurring at different stages of a task~\citep{ji2026prmjudge}. We therefore use a staged progress score, where each rollout receives the score associated with the furthest completion stage reached. Each task is scored on a 10-point scale: the single-goal task uses stages worth 3, 6, and 10 points, while the long-horizon tasks use stages worth 2, 5, 7, and 10 points. Figure~\ref{fig:rubric} illustrates the target state of each stage. We conduct 10 rollouts per task under each evaluation protocol, resulting in 80 rollouts in total, and report the mean score as a percentage of the maximum possible score.

\paragraph{Real-World Baseline Training.}
Since public results are unavailable for our real-robot platform, all real-world baselines are trained using the same demonstrations, observation and action spaces, and training budget as DIDO. Each baseline is trained separately for each task using its released implementation and recommended hyperparameters. $\pi_{0.5}$~\citep{intelligence2025pi05} is initialized from its official pretrained weights, while Fast-WAM~\citep{yuan2026fastwam} is trained on our robot demonstrations. This ensures a consistent data and evaluation setting across methods.

\section{Supplementary Experiments}
\label{app:supp_exp}

This section provides additional task-level results and analyses supporting the main experiments.

\subsection{Effect of Co-Video Supervision}
\label{app:covideo}

Table~\ref{tab:covideo} reports the co-video supervision ablations for Fast-WAM and DIDO. The No co-video variants omit the auxiliary video-generation objective during policy learning. We retain the suite-level and clean/randomized results here to complement the full-model comparisons in Table~\ref{tab:standard}.

\begin{table*}[!t]
\centering
\small
\setlength{\tabcolsep}{1pt}
\renewcommand{\arraystretch}{1.0}
\caption{Effect of co-video supervision on LIBERO and RoboTwin. Results are success rates (\%). All variants are trained without embodied pretraining.}
\label{tab:covideo}
\begin{tabular*}{\textwidth}{@{\extracolsep{\fill}}lcccccccc@{}}
\toprule
& \multicolumn{5}{c}{\textbf{LIBERO}} & \multicolumn{3}{c}{\textbf{RoboTwin}} \\
\cmidrule(lr){2-6} \cmidrule(l){7-9}
Method & Spatial & Object & Goal & Long & Avg. & Clean & Rand. & Avg. \\
\midrule
Fast-WAM (No co-video)
& 89.2 & 99.2 & 95.4 & 90.0 & 93.5 & 82.8 & 84.8 & 83.8 \\
Fast-WAM~\citep{yuan2026fastwam}
& 98.2 & 100.0 & 97.0 & 95.2 & 97.6 & 91.9 & 91.8 & 91.8 \\
\midrule
DIDO (No co-video)
& 97.8 & 99.4 & 97.2 & 96.4 & 97.7 & 89.6 & 89.0 & 89.3 \\
DIDO
& 99.8 & 99.6 & 98.6 & 98.0 & 99.0 & 91.7 & 92.0 & 91.9 \\
\bottomrule
\end{tabular*}
\end{table*}

\subsection{Per-Task RoboTwin Results}
\label{app:robotwin}

\begin{table*}[!t]
\centering
\caption{
Per-task success rates (\%) on RoboTwin.
We compare DIDO without co-video supervision in Stage~\stageII\ with the full model under clean and domain-randomized settings. Each result is averaged over 50 evaluation episodes, with the better result shown in \textbf{bold}.
}
\label{tab:robotwin_pertask}
\footnotesize
\setlength{\tabcolsep}{4pt}
\renewcommand{\arraystretch}{0.95}
\begin{tabular}{@{}lcccc@{}}
\toprule
& \multicolumn{2}{c}{DIDO (No co-video)} & \multicolumn{2}{c}{DIDO} \\
\cmidrule(lr){2-3}\cmidrule(l){4-5}
Task & Clean & Rand. & Clean & Rand. \\
\midrule
\texttt{adjust\_bottle}             & 100 & 98  & 100 & 100 \\
\texttt{beat\_block\_hammer}        & 98  & 98  & 96  & 96  \\
\texttt{blocks\_ranking\_rgb}       & 98  & 98  & 100 & 98  \\
\texttt{blocks\_ranking\_size}      & 78  & 78  & 78  & 78  \\
\texttt{click\_alarmclock}          & 100 & 100 & 98  & 100 \\
\texttt{click\_bell}                & 100 & 100 & 100 & 100 \\
\texttt{dump\_bin\_bigbin}          & 96  & 98  & 96  & 92  \\
\texttt{grab\_roller}               & 100 & 100 & 100 & 98  \\
\texttt{handover\_block}            & 82  & 74  & 94  & 90  \\
\texttt{handover\_mic}              & 98  & 98  & 98  & 98  \\
\texttt{lift\_pot}                  & 100 & 100 & 100 & 100 \\
\texttt{move\_can\_pot}             & 100 & 96  & 72  & 92  \\
\texttt{move\_playingcard\_away}    & 100 & 100 & 100 & 100 \\
\texttt{move\_stapler\_pad}         & 70  & 64  & 90  & 90  \\
\texttt{hanging\_mug}               & 48  & 30  & 50  & 44  \\
\texttt{open\_laptop}               & 98  & 100 & 98  & 98  \\
\texttt{open\_microwave}            & 54  & 50  & 64  & 64  \\
\texttt{pick\_diverse\_bottles}     & 66  & 80  & 82  & 82  \\
\texttt{pick\_dual\_bottles}        & 86  & 86  & 100 & 98  \\
\texttt{place\_a2b\_left}           & 96  & 92  & 98  & 100 \\
\texttt{place\_a2b\_right}          & 94  & 98  & 92  & 96  \\
\texttt{place\_bread\_basket}       & 90  & 88  & 98  & 98  \\
\texttt{place\_bread\_skillet}      & 94  & 88  & 94  & 92  \\
\texttt{place\_can\_basket}         & 82  & 62  & 64  & 60  \\
\texttt{place\_cans\_plasticbox}    & 98  & 100 & 98  & 98  \\
\texttt{place\_container\_plate}    & 100 & 100 & 98  & 98  \\
\texttt{place\_dual\_shoes}         & 78  & 82  & 88  & 84  \\
\texttt{place\_empty\_cup}          & 100 & 100 & 98  & 100 \\
\texttt{place\_fan}                 & 90  & 96  & 98  & 92  \\
\texttt{place\_burger\_fries}       & 94  & 98  & 98  & 100 \\
\texttt{place\_mouse\_pad}          & 92  & 86  & 94  & 98  \\
\texttt{place\_object\_basket}      & 86  & 92  & 88  & 86  \\
\texttt{place\_object\_scale}       & 96  & 90  & 92  & 94  \\
\texttt{place\_object\_stand}       & 96  & 96  & 96  & 98  \\
\texttt{place\_phone\_stand}        & 94  & 92  & 96  & 98  \\
\texttt{move\_pillbottle\_pad}      & 96  & 92  & 100 & 96  \\
\texttt{place\_shoe}                & 90  & 94  & 88  & 94  \\
\texttt{press\_stapler}             & 66  & 94  & 88  & 94  \\
\texttt{put\_bottles\_dustbin}      & 92  & 84  & 96  & 94  \\
\texttt{put\_object\_cabinet}       & 78  & 84  & 88  & 90  \\
\texttt{rotate\_qrcode}             & 86  & 84  & 88  & 86  \\
\texttt{scan\_object}               & 94  & 82  & 94  & 92  \\
\texttt{shake\_bottle}              & 100 & 100 & 100 & 100 \\
\texttt{shake\_bottle\_horizontally}& 100 & 98  & 100 & 100 \\
\texttt{stack\_blocks\_three}       & 94  & 98  & 98  & 98  \\
\texttt{stack\_blocks\_two}         & 100 & 98  & 100 & 98  \\
\texttt{stack\_bowls\_three}        & 86  & 88  & 82  & 86  \\
\texttt{stack\_bowls\_two}          & 96  & 96  & 98  & 96  \\
\texttt{stamp\_seal}                & 84  & 92  & 80  & 86  \\
\texttt{turn\_switch}               & 64  & 60  & 80  & 80  \\
\midrule
\textbf{Average}                    &  89.56 & 89.04 & \textbf{91.72} & \textbf{92.00} \\
\bottomrule
\end{tabular}
\vskip 0.2in
\end{table*}

\begin{figure*}[!t]
\centering
\includegraphics[width=\textwidth]{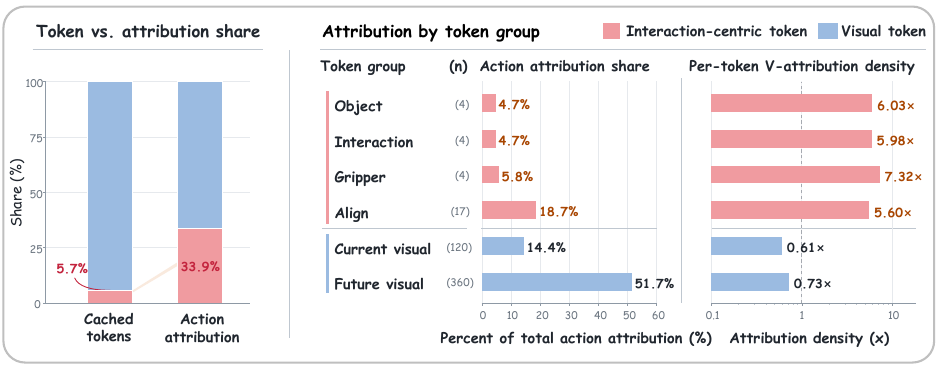}
\caption{
\textbf{Token contributions to action generation.}
\emph{Left}: Interaction-centric tokens occupy only 5.74\% of the cache tokens for action generation but account for 33.9\% of total action attribution.
\emph{Right}: Attribution is further decomposed across token groups. After normalization by token frequency, all interaction-centric groups exhibit substantially higher per-token attribution than the $1\times$ baseline, whereas current and future visual tokens remain below it.
}
\label{fig:prefix_action_attribution}
\end{figure*}

Table~\ref{tab:robotwin_pertask} reports per-task success rates on RoboTwin under both clean and domain-randomized settings. We compare the full DIDO model with a variant that removes the video-generation objective during Stage~II while retaining the action and interaction-centric objectives. Overall, co-video supervision improves the average success rate from 89.56\% to 91.72\% in the clean setting and from 89.04\% to 92.00\% under domain randomization.

The gains are distributed unevenly across individual tasks, which is expected under multi-task joint training, where improvements on some tasks may be accompanied by small degradations on others. Several tasks exhibit particularly large gains, including \texttt{handover\_block}, \texttt{open\_microwave}, \texttt{put\_object\_cabinet}, and \texttt{pick\_dual\_bottles}. Despite occasional task-level regressions, the consistent improvement in overall performance indicates that retaining future-video supervision provides an effective training signal for the multi-task policy.

\subsection{Analysis of Interaction-Centric Tokens}
\label{app:interaction_analysis}

We further investigate whether DIDO's interaction-centric tokens are effectively utilized by the action expert. We refer to all tokens accessible to the action expert as \textbf{cache tokens} and group them by semantic source into \textit{Object}, \textit{Interaction}, \textit{Gripper}, \textit{Alignment}, \textit{Current Visual}, and \textit{Future Visual}. The first four groups constitute the interaction-centric tokens. For each cache token $i$, we quantify its contribution to action generation using the token-level action V-attribution score $a_i$.

\paragraph{Metrics.}
For each token group $g$, we first compute its share of total action attribution as
\begin{equation}
A_g
=
\frac{\sum_{i\in g} a_i}
{\sum_j a_j},
\label{eq:attribution_share}
\end{equation}
where a larger $A_g$ indicates that the group contributes more strongly to action generation. Since groups contain different numbers of tokens, however, $A_g$ alone cannot distinguish token importance from group size. We therefore normalize the attribution share by the token-frequency share and define the per-token attribution enrichment as
\begin{equation}
E_g
=
\frac{A_g}{N_g/N},
\label{eq:attribution_enrichment}
\end{equation}
where $N_g$ is the number of tokens in group $g$ and $N$ is the total number of tokens. $E_g=1$ indicates attribution proportional to token frequency, while $E_g>1$ indicates that tokens in group $g$ contribute more to action generation per token than expected from their frequency. We report means and 95\% confidence intervals using task-level bootstrap with 10K resamples over 50 clips and 31 tasks.

\paragraph{Results.}
Figure~\ref{fig:prefix_action_attribution} shows that action attribution is highly concentrated in the interaction-centric tokens. Although they constitute only 5.74\% of the cache, they account for 33.9\% of total action attribution, corresponding to approximately $5.9\times$ greater attribution per token than expected from their frequency. This indicates that the action expert places disproportionate emphasis on these compact interaction representations.

This pattern is consistent across the individual interaction-centric groups. Object, Interaction, Gripper, and Alignment tokens account for 4.7\%, 4.7\%, 5.8\%, and 18.7\% of total action attribution, respectively, with per-token enrichments of $6.03\times$, $5.98\times$, $7.32\times$, and $5.60\times$. In contrast, Current Visual and Future Visual tokens achieve enrichments of only $0.61\times$ and $0.73\times$, despite comprising most of the cache. These results suggest that the interaction-centric tokens concentrate action-relevant information into a small token budget, complementing the broader scene and future context provided by the visual tokens.

\section{Limitations}
\label{app:limitations}

DIDO has several limitations. First, its interaction-centric reasoning relies on bounding-box supervision for the gripper and target object. Although effective in our settings, such coarse object-centric supervision may be insufficient for multi-object interaction, deformable objects, or fine-grained contact geometry. The required annotations are generated automatically, but detection and tracking failures under severe occlusion or ambiguous object references can still introduce noisy supervision.

Second, DIDO introduces additional training complexity and computational cost. Training involves three separate optimization phases, including robot-domain teacher preparation, one-step distillation, and joint policy learning, making the overall pipeline more involved than directly training an action policy. Finally, the single-step world model is distilled from a robot-adapted four-step teacher, so its predictive capability remains dependent on the quality and coverage of the teacher. Extending DIDO to richer interaction representations, simpler training pipelines, and broader embodiments and environments remains an important direction.

\end{document}